\documentclass[preprint,review,11.5pt]{elsarticle}

\usepackage{hyperref}
\usepackage[usenames,dvipsnames]{xcolor}
\usepackage[framemethod=TikZ]{mdframed}
\usepackage{multicol}
\usepackage{changes}
\usepackage{graphicx}
\usepackage{url}
\usepackage{todonotes}
\usepackage{amsmath}
\usepackage{amssymb}
\usepackage{bm}
\usepackage{natbib}
\usepackage{microtype}
\usepackage{multirow}
\usepackage{paralist}
\usepackage{amsfonts}
\usepackage{enumitem}
\usepackage{tcolorbox}
\usepackage{booktabs}
\usepackage{tabularx}
\usepackage{rotating}
\usepackage{marginnote}
\usepackage{makecell}%% The amsthm package provides extended theorem environments
\usepackage{amsthm}

\usepackage{amsmath}
\usepackage{amsthm}
\usepackage{amssymb}
\usepackage{fancyvrb}

\usepackage[labelsep=period]{caption}

\usepackage{array}
\usepackage{tabu}   % Wide lines in tables
\usepackage{xspace} % Non-eatable spaces in macros

\usepackage{graphicx}
\graphicspath{{figures/}}

\usepackage{placeins}

\usepackage{color}
\usepackage{xcolor}
\definecolor{mauve}{rgb}{0.58,0,0.82}

\usepackage[ruled, vlined, linesnumbered]{algorithm2e}

\SetKw{True}{true}
\SetKw{False}{false}
\SetKwData{typeInt}{Int}
\SetKwData{typeRat}{Rat}
\SetKwData{Defined}{Defined}
\SetKwFunction{parseStatement}{parseStatement}

\usepackage{todonotes}

\usepackage{verbatim}

\usepackage{enumitem}

\usepackage{listings}
\usepackage{proof}
\usepackage{semantic}
\usepackage{declare-macros}
\usepackage{multirow}
\usepackage{epsfig}
\usepackage{subcaption}

\begingroup
\catcode`\~=11
\gdef\urltilde{\lower 0.6ex\hbox{~}}
\endgroup

\newcommand{\Nat}{{\rm I\kern-.23em N}}

\renewcommand{\varsigma}{\mathit{R}}
\newcommand{\LTL}{{\sc ltl}\xspace}

\newcommand{\declare}{{Declare}\xspace}

\newcommand{\Next}{\raisebox{-0.27ex}{\LARGE$\circ$}}

\newcommand{\Until}{\mathop{\mathcal{U}}}

\journal{Decision Support Systems}

\hypersetup{
  bookmarks=true,         % show bookmarks bar?
  unicode=false,          % non-Latin characters in Acrobat’s bookmarks
  pdftoolbar=true,        % show Acrobat’s toolbar?
  pdfmenubar=true,        % show Acrobat’s menu?
  pdffitwindow=false,     % window fit to page when opened
  pdfstartview={FitH},    % fits the width of the page to the window
  pdftitle={},    % title
  pdfauthor={},     % author
  pdfsubject={},   % subject of the document
  pdfcreator={},   % creator of the document
  pdfproducer={}, % producer of the document
  pdfkeywords={keyword1, key2, key3}, % list of keywords
  pdfnewwindow=true,      % links in new PDF window
  colorlinks=true,       % false: boxed links; true: colored links
  linkcolor=red,          % color of internal links
  citecolor=red,        % color of links to bibliography
  filecolor=red,      % color of file links
  urlcolor=Maroon           % color of external links
}

\def\post#1{\ensuremath{{#1}\kern-.05ex-}}

\DeclareCaptionType{copyrightbox}

\RequirePackage{array}
\RequirePackage{tabularx}
\newcolumntype{Y}{>{\centering\let\newline\\\arraybackslash\hspace{0pt}}X}
\newcolumntype{Z}{>{\raggedleft\let\newline\\\arraybackslash\hspace{0pt}}X}
\input{addon/claudio-macros}
\def\paramx {\taskize{A}}
\def\paramy {\taskize{B}}

\newcommand{\taskize}[1] {\ensuremath{\scalebox{0.85}{\textsf{#1}}}}

\def\ExiTxt {Existence}

\def\AbseTxt {Absence}

\def\InitTxt {Init}
\def\EndTxt {End}
\def\ResExTxt {RespondedExistence}
\def\RespTxt {Response}
\def\AltRespTxt {AlternateResponse}

\def\ChaRespTxt {ChainResponse}
\def\PrecTxt {Precedence}
\def\AltPrecTxt {AlternatePrecedence}

\def\ChaPrecTxt {ChainPrecedence}
\def\CoExiTxt {CoExistence}
\def\SuccTxt {Succession}
\def\AltSuccTxt {AlternateSuccession}

\def\ChaSuccTxt {ChainSuccession}
\def\NotCoExiTxt {NotCoExistence}
\def\NotSuccTxt {NotSuccession}
\def\NotChaSuccTxt {NotChainSuccession}
\def\ExiTmp {\ensuremath{\textsc{\ExiTxt}}}

\def\AbseTmp {\ensuremath{\textsc{\AbseTxt}}}

\def\InitTmp {\ensuremath{\textsc{\InitTxt}}}

\def\RespTmp {\ensuremath{\textsc{\RespTxt}}}

\def\PrecTmp {\ensuremath{\textsc{\PrecTxt}}}

\newcommand{\Exi}[2] {\ensuremath{\textsc{\ExiTxt}(#1,#2)}}

\newcommand{\Abse}[2] {\ensuremath{\textsc{\AbseTxt}(#1,#2)}}

\newcommand{\Ini}[1] {\ensuremath{\textsc{\InitTxt}(#1)}}
\newcommand{\End}[1] {\ensuremath{\textsc{\EndTxt}(#1)}}
\newcommand{\ResEx}[2] {\ensuremath{\textsc{\ResExTxt}(#1,#2)}}
\newcommand{\Resp}[2] {\ensuremath{\textsc{\RespTxt}(#1,#2)}}

\newcommand{\AltResp}[2] {\ensuremath{\textsc{\AltRespTxt}(#1,#2)}}

\newcommand{\ChaResp}[2] {\ensuremath{\textsc{\ChaRespTxt}(#1,#2)}}

\newcommand{\Prec}[2] {\ensuremath{{\textsc{\PrecTxt}}(#1,#2)}}
\newcommand{\AltPrec}[2] {\ensuremath{\textsc{\AltPrecTxt}(#1,#2)}}

\newcommand{\ChaPrec}[2] {\ensuremath{\textsc{\ChaPrecTxt}(#1,#2)}}
\newcommand{\CoExi}[2] {\ensuremath{\textsc{\CoExiTxt}(#1,#2)}}
\newcommand{\Succ}[2] {\ensuremath{\textsc{\SuccTxt}(#1,#2)}}
\newcommand{\AltSucc}[2] {\ensuremath{\textsc{\AltSuccTxt}(#1,#2)}}

\newcommand{\ChaSucc}[2] {\ensuremath{\textsc{\ChaSuccTxt}(#1,#2)}}
\newcommand{\NotCoExi}[2] {\ensuremath{\textsc{\NotCoExiTxt}(#1,#2)}}
\newcommand{\NotSucc}[2] {\ensuremath{\textsc{\NotSuccTxt}(#1,#2)}}
\newcommand{\NotChaSucc}[2] {\ensuremath{\textsc{\NotChaSuccTxt}(#1,#2)}} 
\input{addon/commands-for-revision}

\begin{document}

\begin{frontmatter}



\title{Exploring Business Process Deviance with Sequential and Declarative Patterns}

\author[NewCastle]{Giacomo Bergami}
\author[FBK]{Chiara Di Francescomarino}
\author[FBK]{Chiara Ghidini}
\author[Bolzano]{Fabrizio Maria Maggi}
\author[Tartu]{Joonas Puura}

\address[NewCastle]{Newcastle University, Newcastle Upon Tyne, United Kingdom \\
%\email{\{stephan.fahrenkrog-petersen,matthias.weidlich\}@hu-berlin.de}
}
\address[FBK]{Fondazione Bruno Kessler, Trento, Italy \\
%\email{\{stephan.fahrenkrog-petersen,matthias.weidlich\}@hu-berlin.de}
}
\address[Bolzano]{Free University of Bozen-Bolzano, Italy\\
%\email{\{irene.teinemaa,marlon.dumas,f.m.maggi\}@ut.ee}
}
\address[Tartu]{University of Tartu, Estonia\\
%\email{\{irene.teinemaa,marlon.dumas,f.m.maggi\}@ut.ee}
}

\begin{abstract}
Business process deviance refers to the phenomenon whereby a subset of the executions of a business process deviate, in a negative or positive way, with respect to {their} expected or desirable outcomes. Deviant executions of a business process include those that violate compliance rules, or executions that undershoot or exceed performance targets. Deviance mining is concerned with uncovering the reasons for deviant executions by analyzing event logs stored by the systems supporting the execution of a business process. In this paper, the problem of explaining deviations in business processes is first investigated by using features based on sequential and declarative patterns, and a combination of them. Then, the explanations are further improved by leveraging the data attributes of events and traces in event logs through features based on pure data attribute values and data-aware declarative rules. The explanations characterizing the deviances are then extracted by direct and indirect methods for rule induction. Using real-life logs from multiple domains, a range of feature types and different forms of decision rules are evaluated in terms of their ability to accurately discriminate between non-deviant and deviant executions of a process as well as in terms of understandability of the final outcome returned to the users.
\end{abstract}


\end{frontmatter}


\section{Introduction}
\label{sec:intro}

The increasing adoption of ERP (Enterprise Resource Planning) and management applications able to track information about process executions in the so called execution \emph{traces}, has opened up the possibility of extracting knowledge from traces, collected in \emph{event logs}.
Different techniques have been developed in the context of \emph{process mining} to \emph{discover} models from event logs, to \emph{check} conformance between an event log and a process model, or to \emph{enhance} existing process models starting from event logs \cite{DBLP:books/sp/Aalst16,DBLP:journals/tkde/AugustoCDRMMMS19,DBLP:journals/tkdd/OstovarLR20}.
%several process mining techniques, focus on extracting knowledge from event logs.
Among the different types of knowledge that can be extracted from event logs, a crucial role is played by the explanation of \emph{deviant} traces, i.e., business process executions that deviate in a positive or negative way from the expected outcome \cite{DBLP:journals/tkdd/TeinemaaDRM19}. Indeed, discovering why some executions take more (less) time than others, or what characterizes the traces that end up with a faulty (or particularly good) outcome could be very useful for business analysts in order to understand what can be improved to reduce negative deviances and spread the positive ones.

\emph{Business process deviance mining} is a branch of process mining which aims at analyzing event logs in order to discover and characterize business process deviances.
%Deviant executions of a business process include those traces that do not reach or achieve targets (e.g., very slow and very fast traces), or those that violate some constraints.
%violate compliance rules, or executions that undershoot or exceed performance targets. %Deviance mining is concerned with uncovering the reasons for deviant executions by analyzing business process event logs.
The input of deviance mining approaches is an event log, in which each trace is labeled as \emph{deviant} or \emph{non-deviant}. The purpose is to discover a descriptive and informative \emph{model} distinguishing the ``good'' traces from the deviant ones. A good characterization of deviant executions gives analysts hints concerning the causes generating deviance within trace executions, thus allowing effective process improvement solutions.

As the former formulation easily boils down to a binary classification problem, we show how business process deviance mining can be solved following this intuition. Relevant patterns and/or data attributes describing execution traces are used as features for encoding labeled traces. The labeled encoded traces are then used for training a classifier that is in charge of discriminating between deviant and non-deviant executions based on those patterns.
The most obvious choice to describe execution traces~\cite{6597225}, which are sequences of activities, is resorting to \emph{sequential} patterns representing sequences of adjacent activities. However, other types of features can be used to describe a process trace, as for instance \emph{declarative} patterns~\cite{pesic2008}, i.e., patterns related to the validity of predefined temporal properties of activities in the trace, or combinations of sequential and declarative patterns (\emph{hybrid encoding}). In addition to features extracted from the control-flow perspective (expressing properties of the sequence of activities in a trace), it is also possible to extract additional features by making use of data attributes attached to traces and events in an event log, which can help in characterizing differences between deviant and non-deviant executions.

This paper frames the problem of investigating the impact that different types of control-flow features (sequential and declarative patterns, and their combinations) have on business process deviance mining results. Then, it provides two different ways of using data features in addition to control-flow features to characterize business process deviance. These features are evaluated in terms of their ability to accurately discriminate between non-deviant and deviant executions of a process using real-life event logs from multiple domains. Finally, the paper analyzes the possible outcomes returned to the users. Two different methods returning decision rules are compared both in terms of their classification performance and in terms of amount and length of the decision rules returned (to investigate user readability and explanation conciseness). More concretely, the two methods leverage, respectively, decision trees (and a procedure for the extraction of decision rules from them), and the Ripper-$k$ algorithm, which inherently provides decision rules as outcome. The comparison is conducted to investigate the trade-off between the accuracy of the deviance mining approach and the complexity of the decision rules returned.

The paper is structured as follows. Section~\ref{sec:related_work} gives an overview of previous research related to this paper. Section~\ref{sec:background} introduces the necessary background knowledge to understand the concepts and techniques used. In Section~\ref{sec:problem}, a problem statement for business process deviance mining and a motivating example are given. Section~\ref{sec:allapproaches} introduces the pipeline used for business process deviance mining and how sequential and declarative control-flow patterns as well as data attributes can be used as features to support deviance mining. This section also shows how to provide explanations for business process deviance in the form of decision rules. Section~\ref{sec:allevaluations} gives an overview on how the evaluation was carried out and reports the results.
%In particular, Section~\ref{eval:first} reports procedure and results of the experiments evaluating the impact of different types of features based on control-flow. Section~\ref{eval:second} describes procedure and results of the experiments evaluating the impact of the payload-based features. Section~\ref{eval:third} discusses the experiments related to the accuracy/understandability of decision rules returned by the proposed approach.
Finally, Section~\ref{conclusions} concludes the paper by giving an overview of the work done and spells out directions for future work. 
\section{Related Work}
\label{sec:related_work}
The main works related to business process deviance mining can be classified into two main families: the ones using delta-analysis that are mainly based on the identification of differences between the models discovered from deviant and non-deviant traces (e.g., \cite{Suriadi2014,DBLP:conf/bpm/Armas-CervantesBDG14}), and those based on classification techniques~\cite{Suriadi2013,Partington2015,bose2009,6597225,Lo2007,Chen2011}. This work falls in the latter group. The works in this second group leverage classification techniques to discriminate between normal and deviant traces. These approaches usually discover patterns that are then used to build a classifier. They can be further classified based on the type of features used for training the classifier.

In~\cite{Suriadi2013,Partington2015}, the authors use the frequency of individual activities in order to train classifiers in a financial and a clinical scenario, respectively. Bose and van der Aalst in~\cite{bose2009,6597225} employ sequential pattern mining to discover sequential patterns as tandem repeats, maximal repeats and alphabet repeats to be used as features for training a classifier. Similarly, in~\cite{Lakshmanan2013}, association rules are used to discover co-occurrence patterns in the context of deviant classes in a healthcare scenario. In~\cite{Lo2007,Chen2011}, discriminative mining is used to discover discriminative patterns, i.e., patterns that, although not necessarily very frequent, clearly discriminate between deviant and non-deviant traces.
%Differently from these works that propose new patterns to be used as features, the aim of this paper is investigating the impact of different types of features.
A benchmark collecting all these works and evaluating and comparing them in terms of different feature types and classifiers is presented in~\cite{Nguyen2014,Nguyen2016}. %The focus of these works, however, is on the individual, sequential and discriminative patterns separately. In this paper we evaluate the impact of a combination of two families of patterns.

Different types of patterns have also been combined together in~\cite{Cuzzocrea2015,Cuzzocrea2016,Cuzzocrea2017,DBLP:journals/ijcis/CuzzocreaFGP16}. In particular, in~\cite{Cuzzocrea2015}, in order to avoid the redundant representation deriving from mixing different families of patterns, the authors propose an ensemble learning approach in which multiple learners are trained encoding the log according to different types of patterns. In~\cite{Cuzzocrea2016}, data attributes have also been taken into account in the discovery phase as well as for training the classifier. Finally, in~\cite{Cuzzocrea2017,DBLP:journals/ijcis/CuzzocreaFGP16}, the authors enhance their previous work~\cite{Cuzzocrea2015} by proposing an alternative multi-learning approach probabilistically combining various classification methods.
%The focus of the works discussed so far, however, is on the individual, sequential and discriminative patterns separately or on the combination of families of behavioral sequential patterns.

%In this paper, the focus is on the discovery of discriminating data-aware {\declare} constraints for deviance mining.
The extraction of data-aware {\declare} constraints from event logs has been previously discussed in \cite{DBLP:conf/bpm/LenoDM18,DBLP:journals/is/LenoDMRP20}. In that work, the focus is on the unsupervised discovery of data-aware {\declare} models. Our work is also related to papers that present approaches for the supervised discovery of declarative models from positive and negative traces like \cite{DBLP:conf/bpm/LammaMMRS07,DBLP:journals/corr/abs-2109-14883,DBLP:conf/bpm/SlaatsDB21}. Recently, declarative patterns have been used to characterize different process variants like in the approaches presented in \cite{DBLP:conf/bpm/CecconiAC21,DBLP:journals/eswa/RichettiJBC22}.

Differently from the related work described above, this paper:
\begin{itemize}[noitemsep]
 \item takes into account a completely different and unexplored hybrid encoding to support deviance mining; 	
 \item combines the family of declarative patterns with sequential patterns by facing the feature redundancy problem with feature selection approaches;
 \item evaluates the use of data features in combination with control-flow features and provides two different methods for data feature extraction from event logs;
 \item compares and evaluates two methods for extracting explanations for business process deviance in the form of decision rules.
\end{itemize}

\section{Background}
\label{sec:background}
This section gives the background information needed to understand the content of the paper.

\subsection{Business Processes and Logs}
As the work in this paper concentrates on business process event logs, we give here an overview of what a business process is and how its data representation looks like.

\subsubsection{Business Process}
A business process is a set of activities, which are performed in order to achieve a particular goal in a business operation.

Some common types of business processes include:
\begin{itemize}[noitemsep]
\item Application-to-approval, where a sequence of activities are executed after the arrival of an application with the end goal of approving or rejecting it. Examples of this type of processes are the application process of students in universities or the hiring process of employees in companies. Typical activities here are ``call the referees'' or ``score the application'';
\item Order-to-cash, which begins with a customer asking for purchasing a product (or using a service) and ends with the product being delivered and the payment for the product being received by the seller. Typical activities here include ``check the inventory for stock'' or ``estimate the price quote for the customer''.
\end{itemize}

While business processes are executed, it is possible to track and store the process execution information for analysis by turning activities and events into execution traces, which are then collected in the form of event logs.

\label{bg:businessprocess}
\subsubsection{Event log}
\label{bg:xlog}
The standard for storage and manipulation of event logs is \textit{XES (eXtensible Event Stream)} \cite{xesstandard}, which is an XML-based format specialized for the storage of event log data.
%\begin{figure}[!htb]
%        \centering
%                \includegraphics[width=0.7\textwidth]{figures/umlxes}
 %       \caption{Meta-model for the XES standard \cite{xesstandard}}
 %       \label{fig:xesstandard}
%\end{figure}
%Figure \ref{fig:xesstandard} shows the UML \cite{Rumbaugh2004} diagram for the complete meta-model of XES standard.
The basic hierarchy of a XES document contains a single log object. The log object can contain any number of trace objects. Each trace can contain any number of event objects.

All event information related to a specific process is contained within a \emph{log}. Some examples of processes could be:
\begin{itemize}[noitemsep]
\item A medical assessment process;
\item A hiring process of employees.
\end{itemize}

A \emph{trace} describes a time-ordered execution of a specific process. Given the above processes, the corresponding traces could be:
\begin{itemize}[noitemsep]
\item The specific assessment of a patient;
\item The specific hiring of an employee in a company.
\end{itemize}

An \emph{event} represents an observed activity at atomic-level. Possible events in the traces given above could be respectively:
\begin{itemize}[noitemsep]
\item The addition of blood test results to the patient's health record;
\item The decision of hiring by a human resource specialist.
\end{itemize}

\paragraph{Attributes.} Log, trace and event objects define the structure of a XES document. The relevant information is stored in attributes describing either the whole log, or a single trace, or a specific event within a trace. According to the standard, there are 6 elementary attribute types: \textit{String}, \textit{Date}, \textit{Int}, \textit{Float}, \textit{Boolean}, \textit{ID}. Additionally, the standard describes two collection type attributes: \textit{List} and \textit{Container}. %XES does not specify a set of required attributes, but provides extensions for introducing commonly understood attributes.
%\btext{Attributes include (but are not limited to) concept, life-cycle, time and organization: those are particularly crucial for many event log analysis techniques, and are then included as standard extensions of XES.}
%Several concepts are shown in example in Figure~\ref{fig:examplelog}, which depicts the first events of a trace from log \textit{Sepsis} (described later in Section \ref{sssec:datasets}). In the example
In particular, a trace has always at least an attribute of type \textit{String} with key \textit{concept:name}, which describes the trace ID. Each event has always at least an attribute \textit{concept:name} representing the name of the activity executed, an attribute \textit{time:timestamp} indicating the time when the event occurred, and an attribute \textit{lifecycle:transition}, which represents the transactional state of the activity (e.g., \emph{start} indicating that the activity has started, or \emph{complete} representing the completion of the activity). Traces might also have a specific attribute attribute (\emph{label}) specifying whether the trace is considered to be deviant or not. Other attributes can be attached to events such as \textit{org:group} describing which group/resource executed the activity. These additional attributes are generically referred to as \emph{event payload}.

%\begin{figure}[!htb]
%        \centering
%                \includegraphics[width=0.7\textwidth]{figures/examplelog}
%        \caption{An example of a portion of XES log.}
%        \label{fig:examplelog}
%\end{figure}

%In this work OpenXES was used, which is an open source reference implementation of First XES standard.

\subsection{Log patterns}
We give now an overview of the relevant pattern types, which can be used to describe log traces.

\subsubsection{Sequential Patterns}

\label{sssec:sequential}

\emph{Sequential patterns}~\cite{6597225} represent one of the pattern types that can be used to describe traces. Sequential patterns are sequences of events that occur frequently in a trace, thus capturing particular control-flow relations in it. Among the main types of sequential patterns, we can find:
\begin{itemize}[noitemsep]
\small
\item \emph{Tandem Repeats (TR)}: this type of pattern denotes sequences of events that are repeated consecutively within a trace; these sequences correspond to process loops.
\item \emph{Maximal Repeats (MR)}: this type of pattern denotes maximal sequences of events that are repeated in an event log; these sequences correspond to sub-processes.
\item \emph{Tandem Repeats Alphabet  (TRA)}: this type of pattern denotes tandem repeats that share the same activities (i.e., the alphabet of unique activities); these sequences correspond to variations of TR taking into account process parallelism.
\item \emph{Maximal Repeats Alphabet (MRA)}: this type of pattern denotes maximal repeats that share the same activities (i.e., the alphabet of unique activities); these sequences correspond to variations of MR taking into account process parallelism.
\end{itemize}

For instance, given a trace $T=\left\langle \activity{a},\activity{b},\activity{c},\activity{a},\activity{b},\activity{c},\activity{d},\activity{a},\activity{b}\right\rangle$, the set of TR is $\{abc\}$, as $abc$ is the only pattern repeated twice consecutively.
For TRA, the ordering of activities within the pattern does not matter, but the patterns have to appear consecutively. In this case, given a trace $T= \left\langle \activity{a},\activity{b},\activity{c},\activity{c},\activity{b},\activity{a},\activity{a},\activity{b},\activity{c}\right\rangle$, the set of TRA is $\{abc, cb, ab, c, a\}$, with $abc$ being repeated 3 times, $ab$ and $cb$ twice and $a$ and $c$ also twice (without a specific order of events within the pattern).
A pattern is considered to be a maximal repeat, if it cannot be extended to left or to right for a longer repeat covering all the occurrences of the pattern. For example, considering trace $T=\left\langle \activity{a},\activity{b},\activity{c},\activity{a},\activity{b},\activity{c},\activity{d},\activity{a},\activity{b}\right\rangle$, the set of MR is $\{ab,abc\}$ ($c$ and $bc$ are not maximal, because both occurrences can be extended to the left to be $abc$, which includes all $c$ and $bc$ in the trace). Pattern $ab$ is a maximal repeat, because extending it would not cover the last occurrence of $ab$.
For MRA, the ordering of activities within the pattern does not matter. For trace $T=\left\langle \activity{b},\activity{a},\activity{c},\activity{a},\activity{b},\activity{c},\activity{d},\activity{b},\activity{a}\right\rangle$, the set of MRA is $\{ab,abc\}$. Pattern $abc$ occurs twice and is therefore a repeat. Pattern $ab$ is repeated three times and is maximal, because it cannot be extended in a way that it covers all the occurrences of $ab$. Patterns $c$, $b$ are not maximal, because all $c$ also occur in $abc$, and all $b$ occur in $ab$, which are both maximal patterns.
%MR and MRA in an event log are found by concatenating all traces in the event log in a single trace.

\subsubsection{Declare}
\label{bg:declare}

{\declare} is a declarative process modeling language first introduced in~\cite{pesic2008}. The declarative approach for business process modeling was introduced to be able to model loosely-structured processes \cite{4384001} working in contexts with high variability. As shown in \cite{DBLP:conf/bpm/PichlerWZPMR11}, this type of processes are indeed difficult to be defined with the rigid specifications of imperative approaches, which tell users what to do step by step. They can instead easily be designed using declarative process models that shift the decision making from the workflow system supporting the process execution to the user. The basic building block of a {\declare} model is a \textit{constraint}, which is a \emph{template} (i.e., an abstract parameterized property) instantiated on a set of real activities.

\begin{table}[t!]
  \centering
  %!TEX root = ../main.tex

\usetikzlibrary{arrows.meta,decorations.markings}
\def\DTZU {2ex}
\tikzset{
 DECLARE.task/.style={
  rounded corners=1,
  minimum width=3em,
  minimum height=1ex,
  draw
 },
 DECLARE.existcon/.style={
  label={[style=draw,yshift=-1\pgflinewidth]above:{\tiny \textit{#1}}}
 },
 DECLARE.neg/.style={
  semithick,
  postaction={decorate,decoration={markings,
   mark=at position .5 with {\arrow[xshift=0.15*\DTZU]{Bar[width=1.5*\DTZU]}},
   mark=at position .5 with {\arrow[xshift=-0.15*\DTZU]{Bar[width=1.5*\DTZU]}}
  }}
 },
 DECLARE.resex/.style={
  semithick,
  {Circle[length=1*\DTZU,width=1*\DTZU]}-,
  shorten <=-0.5*\DTZU
 },
 DECLARE.resp/.style={
  semithick,
  {Circle[length=1*\DTZU,width=1*\DTZU]}-{Triangle[length=1*\DTZU,width=1*\DTZU]},
  shorten <=-0.5*\DTZU
 },
 DECLARE.alt.resp/.style={
  semithick,
  {Circle[length=1*\DTZU,width=1*\DTZU]}-{Triangle[length=1*\DTZU,width=1*\DTZU]},
  shorten <=-0.5*\DTZU,
  double distance=0.25*\DTZU
 },
 DECLARE.chn.resp/.style={
  semithick,
  preaction={draw, shorten <= -0.5*\DTZU, double distance=0.25*\DTZU, {Circle[length=1*\DTZU,width=1*\DTZU]}-{Triangle[length=1*\DTZU,width=1*\DTZU]}},
  {-{Triangle[length=1*\DTZU,width=1*\DTZU]}}
 },
 DECLARE.prec/.style={
  semithick,
  {-{Triangle[length=1*\DTZU,width=1*\DTZU]Circle[length=1*\DTZU,width=1*\DTZU]}},
  shorten >=-0.5*\DTZU
 },
 DECLARE.alt.prec/.style={
  semithick,
  postaction={draw, shorten >= 0.5*\DTZU, double distance=0.25*\DTZU, {-{Triangle[length=1*\DTZU,width=1*\DTZU]}}},
  {-{Circle[length=1*\DTZU,width=1*\DTZU]}},
  shorten >= -0.5*\DTZU
 },
 DECLARE.chn.prec/.style={
  semithick,
  preaction={draw, shorten >= 0.5*\DTZU, double distance=0.25*\DTZU, {-{Triangle[length=1*\DTZU,width=1*\DTZU]}}},
  {-{Circle[length=1*\DTZU,width=1*\DTZU]}},
  shorten >= -0.5*\DTZU
 },
 DECLARE.succ/.style={
  semithick,
  {Circle[length=1*\DTZU,width=1*\DTZU]}-{Triangle[length=1*\DTZU,width=1*\DTZU]Circle[length=1*\DTZU,width=1*\DTZU]},
  shorten <=-0.5*\DTZU,
  shorten >=-0.5*\DTZU
 },
 DECLARE.coex/.style={
  semithick,
  {Circle[length=1*\DTZU,width=1*\DTZU]}-{Circle[length=1*\DTZU,width=1*\DTZU]},
  shorten <=-0.5*\DTZU,
  shorten >=-0.5*\DTZU
 },
 DECLARE.alt.succ/.style={
  semithick,
  postaction={draw, shorten >= 0.5*\DTZU, shorten <= -0.5*\DTZU, double distance=0.25*\DTZU, {Circle[length=1*\DTZU,width=1*\DTZU]}-{Triangle[length=1*\DTZU,width=1*\DTZU]}},
  {-{Circle[length=1*\DTZU,width=1*\DTZU]}},
  shorten >= -0.5*\DTZU
 },
 DECLARE.chn.succ/.style={
  semithick,
  preaction={draw, shorten >= 0.5*\DTZU, shorten <= -0.5*\DTZU, double distance=0.25*\DTZU, {Circle[length=1*\DTZU,width=1*\DTZU]}-{Triangle[length=1*\DTZU,width=1*\DTZU]}},
  {-{Circle[length=1*\DTZU,width=1*\DTZU]}},
  shorten >= -0.5*\DTZU
 }
} 

\renewcommand{\arraystretch}{1.2}
\begin{scriptsize}
\begin{tabular}{ l p{5cm} c}
\toprule
\textbf{Template} & \textbf{Explanation} & \textbf{Notation} \\  % bcaaccbbbaba
\midrule
\multicolumn{3}{l}{Existence templates}\\
\midrule
$\Exi{n}{\paramx}$ & % [^$2]*($2[^$2]*){$1,}[^$2]*
$\paramx$ occurs at least $n$ times &
% & {\paramx} & --<w
\begin{tikzpicture}[baseline=(current bounding box.center)]\node[DECLARE.task,DECLARE.existcon=$n..\ast$]{\paramx};\end{tikzpicture}
\\
$\Abse{m+1}{\paramx}$ & % [^$2]*($2[^$2]*){0,$max}[^$2]*
$\paramx$ occurs at most $m$ times &
% & {\paramx} & --
\begin{tikzpicture}[baseline=(current bounding box.center)]\node[DECLARE.task,DECLARE.existcon=$0..m$]{\paramx};\end{tikzpicture}
\\
$\Ini{\paramx}$ &
{\paramx} is the \emph{first} to occur &
%\taskize{\textbf{\uline{a}}ccbbbaba}
% & {\paramx} & --
\begin{tikzpicture}[baseline=(current bounding box.center)]\node[DECLARE.task,DECLARE.existcon=Init]{\paramx};\end{tikzpicture}
\\
$\End{\paramx}$ &
{\paramx} is the \emph{last} to occur &
%\taskize{bcaaccbbbab\textbf{\uline{a}}}
% & {\paramx} & --
\begin{tikzpicture}[baseline=(current bounding box.center)]\node[DECLARE.task,DECLARE.existcon=End]{\paramx};\end{tikzpicture}
\\
\midrule
\multicolumn{3}{l}{Relation templates}\\
\midrule
$\ResEx{\paramx}{\paramy}$ &
If {\paramx} occurs, then {\paramy} occurs &
%\taskize{\textbf{b}c\textbf{\uline{a}}\uline{a}ccbbb\uline{a}b\uline{a}}
% & {\paramx} & \paramy
\begin{tikzpicture}[baseline=(current bounding box.center),node distance=8*\DTZU]
 \node[DECLARE.task,] (1) {$\paramx$};
 \node[DECLARE.task,right of=1] (2) {$\paramy$};
   
 \path (1) edge [DECLARE.resex] node {} (2);
\end{tikzpicture}
\\
$\Resp{\paramx}{\paramy}$ &
If {\paramx} occurs, then {\paramy} occurs after {\paramx} &
%\taskize{bc\textbf{\uline{a}}\uline{a}cc\textbf{b}bb\textbf{\uline{a}}\textbf{b}}
% & {\paramx} & \paramy
\begin{tikzpicture}[baseline=(current bounding box.center),node distance=8*\DTZU]
 \node[DECLARE.task,] (1) {$\paramx$};
 \node[DECLARE.task,right of=1] (2) {$\paramy$};
   
 \path (1) edge [DECLARE.resp] node {} (2);
\end{tikzpicture}
\\
$\AltResp{\paramx}{\paramy}$ &
Each time {\paramx} occurs, then {\paramy} occurs afterwards, before {\paramx} recurs &
%\taskize{bc\textbf{\uline{a}}cc\textbf{b}bb\textbf{\uline{a}}\textbf{b}}
% & {\paramx} & \paramy
\begin{tikzpicture}[baseline=(current bounding box.center),node distance=8*\DTZU]
 \node[DECLARE.task,] (1) {$\paramx$};
 \node[DECLARE.task,right of=1] (2) {$\paramy$};
   
 \path (1) edge [DECLARE.alt.resp] node {} (2);
\end{tikzpicture}
\\
$\ChaResp{\paramx}{\paramy}$ &
Each time {\paramx} occurs, then {\paramy} occurs immediately after &
%\taskize{bc\textbf{\uline{a}}\textbf{b}bb\textbf{\uline{a}}\textbf{b}}
% & {\paramx} & \paramy
\begin{tikzpicture}[baseline=(current bounding box.center),node distance=8*\DTZU]
 \node[DECLARE.task,] (1) {$\paramx$};
 \node[DECLARE.task,right of=1] (2) {$\paramy$};
   
 \path (1) edge [DECLARE.chn.resp] node {} (2);
\end{tikzpicture}
\\
$\Prec{\paramx}{\paramy}$ &
{\paramy} occurs only if preceded by {\paramx} &
%\taskize{c\textbf{a}acc\textbf{\uline{b}}\uline{b}\uline{b}a\uline{b}a}
% & {\paramy} & \paramx
\begin{tikzpicture}[baseline=(current bounding box.center),node distance=8*\DTZU]
 \node[DECLARE.task,] (1) {$\paramx$};
 \node[DECLARE.task,right of=1] (2) {$\paramy$};
   
 \path (1) edge [DECLARE.prec] node {} (2);
\end{tikzpicture}
\\
$\AltPrec{\paramx}{\paramy}$ &
Each time {\paramy} occurs, it is preceded by {\paramx} and no other {\paramy} can recur in between &
%\taskize{ca\textbf{a}cc\textbf{\uline{b}}\textbf{a}\textbf{\uline{b}}a}
% & {\paramy} & \paramx
\begin{tikzpicture}[baseline=(current bounding box.center),node distance=8*\DTZU]
 \node[DECLARE.task,] (1) {$\paramx$};
 \node[DECLARE.task,right of=1] (2) {$\paramy$};
   
 \path (1) edge [DECLARE.alt.prec] node {} (2);
\end{tikzpicture}
\\
$\ChaPrec{\paramx}{\paramy}$ &
Each time {\paramy} occurs, then {\paramx} occurs immediately before &
%\taskize{c\textbf{a}\textbf{\uline{b}}\textbf{a}\textbf{\uline{b}}a}
% & {\paramy} & \paramx
\begin{tikzpicture}[baseline=(current bounding box.center),node distance=8*\DTZU]
 \node[DECLARE.task,] (1) {$\paramx$};
 \node[DECLARE.task,right of=1] (2) {$\paramy$};
   
 \path (1) edge [DECLARE.chn.prec] node {} (2);
\end{tikzpicture}
\\
\midrule
\multicolumn{3}{l}{Mutual relation templates}\\
\midrule
$\CoExi{\paramx}{\paramy}$ &
If {\paramy} occurs, then {\paramx} occurs, and vice versa &
%\taskize{\textbf{\uline{b}}c\textbf{\uline{a}}cc\uline{bbbaba}}
% & \paramx, {\paramy} & \paramy, \paramx
\begin{tikzpicture}[baseline=(current bounding box.center),node distance=8*\DTZU]
 \node[DECLARE.task,] (1) {$\paramx$};
 \node[DECLARE.task,right of=1] (2) {$\paramy$};
   
 \path (1) edge [DECLARE.coex] node {} (2);
\end{tikzpicture}
\\
$\Succ{\paramx}{\paramy}$ &
{\paramx} occurs if and only if {\paramy} occurs after {\paramx} &
%\taskize{c\textbf{\uline{a}}\uline{a}cc\textbf{\uline{b}}\uline{bbab}}
% & \paramx, {\paramy} & \paramy, \paramx
\begin{tikzpicture}[baseline=(current bounding box.center),node distance=8*\DTZU]
 \node[DECLARE.task,] (1) {$\paramx$};
 \node[DECLARE.task,right of=1] (2) {$\paramy$};
   
 \path (1) edge [DECLARE.succ] node {} (2);
\end{tikzpicture}
\\
$\AltSucc{\paramx}{\paramy}$ &
{\paramx} and {\paramy} occur if and only if the latter follows the former, and they alternate each other &
%\taskize{c\textbf{\uline{a}}cc\textbf{\uline{b}}\textbf{\uline{a}}\textbf{\uline{b}}}
% & \paramx, {\paramy} & \paramy, \paramx
\begin{tikzpicture}[baseline=(current bounding box.center),node distance=8*\DTZU]
 \node[DECLARE.task,] (1) {$\paramx$};
 \node[DECLARE.task,right of=1] (2) {$\paramy$};
   
 \path (1) edge [DECLARE.alt.succ] node {} (2);
\end{tikzpicture}
\\
$\ChaSucc{\paramx}{\paramy}$ &
{\paramx} and {\paramy} occur if and only if the latter immediately follows the former &
%\taskize{c\textbf{\uline{a}}\textbf{\uline{b}}\textbf{\uline{a}}\textbf{\uline{b}}}
% & \paramx, {\paramy} & \paramy, \paramx
\begin{tikzpicture}[baseline=(current bounding box.center),node distance=8*\DTZU]
 \node[DECLARE.task,] (1) {$\paramx$};
 \node[DECLARE.task,right of=1] (2) {$\paramy$};
   
 \path (1) edge [DECLARE.chn.succ] node {} (2);
\end{tikzpicture}
\\
\midrule
\multicolumn{3}{l}{Negative relation templates}\\
\midrule
$\NotCoExi{\paramx}{\paramy}$ &
{\paramx} and {\paramy} never occur together &
%\taskize{c\uline{aa}cc\uline{a}}
% & \paramx, {\paramy} & \paramy, \paramx
\begin{tikzpicture}[baseline=(current bounding box.center),node distance=8*\DTZU]
 \node[DECLARE.task,] (1) {$\paramx$};
 \node[DECLARE.task,right of=1] (2) {$\paramy$};
   
 \path (1) edge [DECLARE.coex,DECLARE.neg] node {} (2);
\end{tikzpicture}
\\
$\NotSucc{\paramx}{\paramy}$ &
{\paramx} never occurs before {\paramy} &
%\taskize{bc\textbf{\uline{a}}\uline{a}cc\uline{a}}
% & \paramx, {\paramy} & \paramy, \paramx
\begin{tikzpicture}[baseline=(current bounding box.center),node distance=8*\DTZU]
 \node[DECLARE.task,] (1) {$\paramx$};
 \node[DECLARE.task,right of=1] (2) {$\paramy$};
   
 \path (1) edge [DECLARE.succ,DECLARE.neg] node {} (2);
\end{tikzpicture}
\\
$\NotChaSucc{\paramx}{\paramy}$ &
{\paramx} and {\paramy} occur if and only if the latter does not immediately follow the former &
%\taskize{bc\uline{a}\textbf{\uline{a}}cc\textbf{b}bbb\uline{a}}
% & \paramx, {\paramy} & \paramy, \paramx
\begin{tikzpicture}[baseline=(current bounding box.center),node distance=8*\DTZU]
 \node[DECLARE.task,] (1) {$\paramx$};
 \node[DECLARE.task,right of=1] (2) {$\paramy$};
   
 \path (1) edge [DECLARE.chn.succ,DECLARE.neg] node {} (2);
\end{tikzpicture}
\\
\bottomrule
\end{tabular}
\end{scriptsize} 
  \caption{{\declare} templates}
  \label{tab:dec}
\end{table}

Linear Temporal Logic (\LTL) over finite traces is used for specifying the formal semantics of {\declare} templates \cite{Montali2010:Choreographies}.
%The \LTL operators used for describing the semantics of the templates are described in Table~\ref{operators}.
In addition, {\declare} templates have a graphical notation, which makes them easy to use and interpret for process analysts. \tablename~\ref{tab:dec} gives an overview of the most commonly used {\declare} templates, their graphical representations, and a textual description for each of them. The parameters of a template are specified in capital letters, whereas real activities of constraints are specified in lower-case letters (e.g., \RespTmp(\activity{a},\activity{b}) is an instantiation of template \RespTmp\ with activities \activity{a} and \activity{b}).

{\declare} templates can be grouped into three main categories: \emph{existence} templates (first 4 rows of \tablename~\ref{tab:dec}), which involve only one event; \emph{(mutual) relation} templates (rows from 5 to 15), which describe a dependency between two events; and \emph{negative relation} templates (last 3 rows), which describe a negative dependency between two events.
To give some examples of {\declare} constraints, consider the following four traces:

\begin{enumerate}[noitemsep]
\item $\langle \activity{a}, \activity{a}, \activity{b}, \activity{c} \rangle$;
\item $\langle \activity{b}, \activity{b}, \activity{c}, \activity{d} \rangle$;
\item $\langle \activity{a}, \activity{b}, \activity{c}, \activity{b} \rangle$;
\item $\langle \activity{a}, \activity{b}, \activity{a}, \activity{c} \rangle$.
\end{enumerate}

Constraint \InitTmp(\activity{a}) (meaning that a trace has to start with the execution of \activity{a}) is satisfied in traces 1, 3 and 4, but not satisfied in trace 2, since this trace does not start with \activity{a}. On the other hand, \RespTmp(\activity{a}, \activity{b}) (meaning that if \activity{a} occurs, then \activity{b} must eventually follow) is satisfied in traces 1, 2 and 3, but not satisfied in trace 4, since, here, the second occurrence of \activity{a} is not eventually followed by \activity{b}.

An \emph{activation} of a constraint in a trace is an event whose occurrence imposes obligations on another event (the \emph{target}) in the context of same trace \cite{DBLP:conf/bpm/MaggiMCM16,DBLP:journals/is/CiccioMMM18}. For example, for constraint \RespTmp(\activity{a},\activity{b}) \activity{a} is an activation, because the execution of $\activity{a}$ imposes an obligation on $\activity{b}$, forcing it to be eventually executed. Event \activity{b} is a target. Referring back to the sample traces above, in traces 1, 3 and 4, \activity{a} occurs and constraint \RespTmp(\activity{a},\activity{b}) is activated. Trace 2 does not include \activity{a} and, therefore, the constraint is not activated in that trace. In such a case, the constraint is \emph{vacuously satisfied}~\cite{conf/charme/KupfermanV99} in the trace.

An activation of a constraint in a trace is either a \emph{fulfillment} or a \emph{violation} for the constraint in the trace. If every activation of a constraint in a trace leads to a fulfillment, then the constraint is \emph{satisfied} in the trace. Constraint \RespTmp(\activity{a},\activity{b}) is activated in traces 1, 3 and 4. In trace 1, it is activated twice and both occurrences of \activity{a} are eventually followed by \activity{b}, therefore both activations are fulfillments. In trace 3, the constraint is activated and fulfilled once. In trace 4, the constraint is activated twice, but the second activation \activity{a} is not eventually followed by \activity{b} and leads to a violation. A constraint is \emph{violated} in a trace if at least one activation of the constraint leads to a violation in the trace.

%\paragraph{Vacuous satisfaction.} There exist cases in which the constraint is not activated at all. Consider, for instance, the sample trace 2 - $\left\langle \activity{b}, \activity{b}, \activity{c}, \activity{d}\right\rangle$. The considered \RespTmp\ constraint is satisfied in a trivial way in this trace, since \activity{a} never occurs.

\subsubsection{Data-Aware Declare Constraints}
\label{bg:dad}

While {\declare} in its original form is mainly used to set constraints on control-flow aspects of a process, data-aware {\declare} constraints extend {\declare} constraints so as to include conditions on data payloads of events \cite{dataawaredeclare}. A data payload of an event is a list of pairs attribute-value including the event attributes (which can have different values for different events in the same trace) and the trace attributes (having the same value for all events in the same trace) In particular, a data-aware {\declare} constraint is activated if its (control-flow based) activation occurs and an additional condition on its data payload holds. Consider, for example, the following traces:

\begin{enumerate}[noitemsep]
\item $\langle \activity{a} \{g=1\}, \activity{a} \{g=2\}, \activity{b}, \activity{c} \rangle$;
\item $\langle \activity{b}, \activity{b}, \activity{c}, \activity{d} \rangle$;
\item $\langle \activity{a} \{g=2\}, \activity{b}, \activity{c}, \activity{b} \rangle$;
\item $\langle \activity{a} \{g=2\}, \activity{b}, \activity{a} \{g=1\}, \activity{c} \rangle$.
\end{enumerate}

In these traces, $\activity{a} \{g=1\}$ means that $\activity{a}$ has an attribute \textit{g} having value \textit{1} in its payload.
Consider the data-aware constraint \RespTmp(\activity{a},\activity{b},\{g=1\}). This constraint requires the occurrence of \activity{a} with attribute $g=1$ to be activated. In trace 1, the first \activity{a} is an activation, but the second one is not, since the data condition does not hold on its payload. The only activation in this trace leads to a fulfillment (since it is eventually followed by \activity{b}) and the constraint is satisfied in that trace. Trace 2 does not have any activations and is, therefore, vacuously satisfied. In trace 3, there are no activations either since $\activity{a}$ occurs only once, but the data condition does not hold on its payload. In trace 4, the constraint is activated once (second occurrence of \activity{a}) and this leads to a violation since the occurrence of the activation is not followed by the occurrence of a target \activity{b}.

\section{Problem} % (fold)
\begin{figure}[!th]
        \centering
                \includegraphics[width=\textwidth]{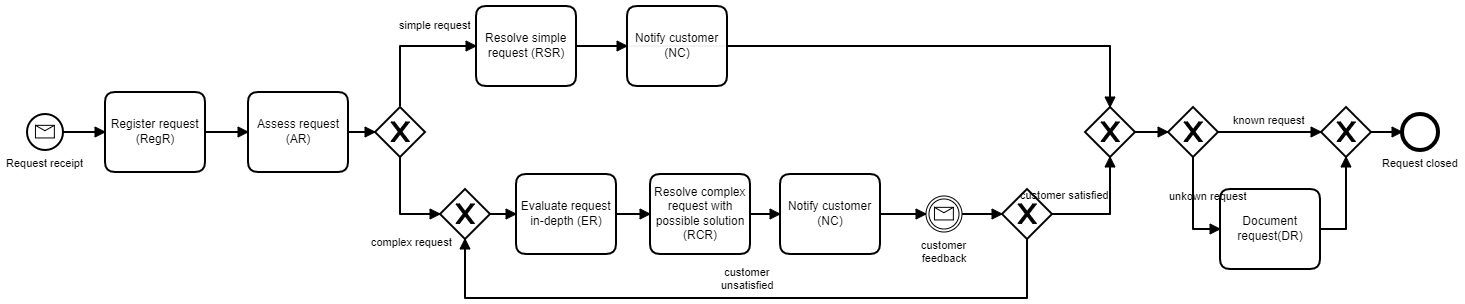}
        \caption{Customer support example}
        \label{fig:bpmn}
\end{figure}

\label{sec:problem}
This section provides an example motivating the importance of approaches for business process deviance mining. Later, we will show the advantages of exploring different types of patterns to explain deviances based on this example (see Section \ref{ssec:application}). The example pertains to the customer support process carried out by a company providing a service, shown in the form of a BPMN model in \figurename~\ref{fig:bpmn}. The support process aims at helping customers whenever a problem occurs or when new features are required. The process starts when the customer support office receives a request from a client. The customer support office registers the request and performs a first evaluation. If the request can be easily solved, it is solved and the customer is notified. In case the request is difficult to solve, a more in-depth evaluation is carried out, a solution is found and the customer is notified. Then the customer is asked to give feedback on the proposed solution. If the customer is not happy with the solution, the issue is evaluated again, an alternative solution is found and the customer is notified again until a suitable solution is found. In both cases (simple and complex requests), if the problem was never encountered before, a report has to be prepared documenting the request and the solution provided before the request can be closed.

 \begin{figure}[!th]
\includegraphics[width=1\textwidth]{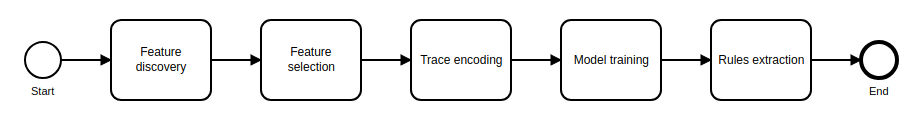}
\centering
\caption{High-level pipeline}
\label{fig:pipeline}
\end{figure}

The company's business process analysts have noticed that some of the executions of the customer support process take longer than others, i.e., these executions \emph{deviate} with respect to the expected process execution time. Analysts are, therefore, interested in understanding what these deviant traces are, how to characterize them and the reasons why they take more time. This information, is indeed crucial for them to be able to improve the process and avoid the occurrence of these delayed executions in the future.

\section{Business Process Deviance Mining Pipeline}
\label{sec:allapproaches}
%This section shows how to use sequential and declarative patterns for explaining deviances in business processes. Then, it shows how to use payload features in combination with control-flow features for deviance mining and provides two different methods for extracting payload features from event logs. Finally, it introduces two methods for providing explanations for business process deviance in the form of decision rules.
%\subsection{Exploring Business Process Deviance with Sequential and Declarative Patterns}
%\label{sec:approach}
%{\color{blue} by taking as a further input the $k$ parameter for the stratified $k$-fold algorithm, and the minimum threshold $\vartheta$ for the feature selection. In addition to that, we can also indicate the most appropriate set of trace encoding strategies $\mathcal{S}$ that we can assess within the pipeline. In particular, the stratified $k$-fold algorithm subdivides the log into $k$ sublogs while guaranteeing an uniform distribution of the classes for each subglog; in particular, we make sure that each log contains at least the $70\%$ of non-deviant traces: then, we run the pipeline in \figurename~\ref{fig:pipeline} for each of the sublogs.}
Given a labeled log, as first step of the pipeline, we carry out \textit{feature discovery} (Section \ref{sec:cfFea}) by exploiting sequential pattern discovery, {\declare} discovery methods, and by using a combination of the two; in addition to control-flow feature, in some cases, also features extracted from event payloads is used. Due to the large number of features that can be discovered, the next step is to perform \textit{feature selection} (Section \ref{declare:featureselection}) by extracting a subset of features that are used to encode the traces.
%In this step, the selected control-flow features can be combined with data features extracted from the log to take the data perspective into consideration.
Then, \textit{trace encoding} (\S\ref{approach:overEnco}) provides a feature vector embedding for each trace; to each vector we associate a label indicating whether the corresponding trace is deviant or not with respect to a certain criterion.
%Please observe that the trace encoding section might introduce some noise, as there could possibly be two distinct traces having different tagging values but having the same feature vectors (\textit{encoding inconsistency}). In order to overcome this limitation,
Then, in the \textit{model training} phase, we exploit classification models for establishing a correlation between the dimensions of the feature vector and the labels. In addition to that, such models will act as explainer substantiating the difference between deviant traces and non-deviant traces.
%furthermore, for each model we exploit a parameter tuning algorithm visiting a subset $\Theta$ of the hyper-parameter space: for each hyper-parameter $\theta\in\Theta$ and for each trace encoding strategy, we assess the model's precision, recall, and AUC (area under the ROC curve). As a last remark, both $k$ and $\vartheta$ parameters could be tuned up to reduce the amount of primary memory required for both the trace encoding and the model training phase.}

\subsection{Feature Discovery}\label{sec:cfFea}
Feature discovery is the process of extracting the relevant  characteristics to all the traces within the log: those could be described as sequential patterns, declarative rules, or specific payload values. These features will be later on represented as one dimension of a vector that will provide the trace representation. Such a vector, alongside with the label of the associated trace, will be the only input for the model training task.

\paragraph*{Data Features}
\label{payload:puredata} Concerning pure data features extracted from the event payloads, we adopt the attribute-value data representation. Such a representation is widely adopted in data mining, machine learning, neural networks, and statistics due to its simplicity. For each attribute $X$ in the payload, we extract a set of values $v\in \mathbf{V}_X$ associated to $X$, so to generate a new data feature using these values. Data features can be of type \textit{String}, \textit{Int}, \textit{Float} and \textit{Boolean}.

As attributes $X$ within a single trace might be associated with different values in different events, we first decided to prioritize either the first occurrence of $X$ (\textit{Choose first}, denoted as $\texttt{first(}X\texttt{)}=v$) or its last occurrence (\textit{Choose last}, denoted as $\texttt{last(}X\texttt{)}=v$). E.g., consider trace $\langle \activity{a} \{g = 1\}, \activity{a} \{g = 2\}, \activity{b} \{g = 3\}, \activity{c} \rangle$. For attribute \textit{g}, \textit{Choose first} and \textit{Choose last} produce features \texttt{first(}\textit{g}\texttt{)}\textit{ = 1} and \texttt{last(}\textit{g}\texttt{)}\textit{ = 3}.%, since these are the values of its first and last occurrence, respectively.

As an alternative strategy, we also decided to compute aggregation functions $f$ over a multiset\footnote{A multiset is a pair $(A,\mu)$, where $A$ is a set of values and $\mu:A\to\mathbb{N}$ denotes the number of occurrences $\mu(a)$ of each $a\in A$.} of values $(\mathbf{V}_X,\mu)$ for each given $X$; we restricted the class of aggregation functions to the following, thus producing a feature denoted as $\texttt{f}(X)=u$, where $u$ is the result of the aggregation function.
\begin{itemize}[noitemsep]
	\item \textbf{Value $v$ Count} ($\texttt{count}(X,v)=\mu(v)$) - For each value $v\in\mathbf{V}_X$ associated to an attribute $X$, we count  how many times ($\mu(v)$) the attribute assumes that value in the trace (used for attributes of type \textit{String});
	\item \textbf{Choose max} ($\texttt{max}(X)=\max\mathbf{V}_X$) - The value of the feature is the maximum value of the attribute in the trace (used for attributes of type \textit{Int} and \textit{Float});
	\item \textbf{Choose min} ($\texttt{min}(X)=\min\mathbf{V}_X$) - The value of the feature is the minimum value of the attribute in the trace (used for attributes of type \textit{Int} and \textit{Float});
	\item \textbf{Compute avg} ($\texttt{avg}(X)=\textup{avg}\;\mathbf{V}_X$) - The value of the feature is the average of the values the attribute assumes in the trace (used for attributes of type \textit{Int} and \textit{Float}).
\end{itemize}
E.g., given a trace $\langle \activity{a} \{g=1\}, \activity{a} \{g=2\}, \activity{b} \{g=3\}, \activity{c} \rangle$, \textit{Choose max}, \textit{Choose min} and \textit{Compute avg} produce features \texttt{max}\textit{(g) = 3}, \texttt{min}\textit{(g) = 1} and \texttt{avg}$(g) = 2$, respectively. Also, given a trace $\langle \activity{a} \{\textit{color}=\textit{white}\}, \activity{a} \{\textit{color}=\textit{black}\}, \activity{b} \{\textit{color}=\textit{white}\}, \activity{c} \rangle$,  attribute \textit{color} assumes two different values: \textit{white} and \textit{black}. Therefore, \textit{Count} produces two features. The first feature is \texttt{count}\textit{(color,\;white) = 2} and the second one is \texttt{count}\textit{(color,\;black) = 1}.
Finally, we also extract meta-information of a trace, such as the \textit{trace length} (the number of events within the trace) and the \textit{trace time length} (time difference in milliseconds between the last and the first event of the trace). All these feature extraction techniques were exploited for our experiments.

\paragraph*{Sequential Features}
%In order to transform traces into feature vectors, the first step is to discover sequential and declarative control-flow patterns.
We discover sequential patterns as in \cite{6597225} to extract discriminative TR, TRA, MR and MRA patterns. This algorithm discovers patterns that are frequent either in deviant traces or in non-deviant traces but not in both, so to identify features that uniquely describe each class of traces. Frequent patterns are selected by filtering out the ones having a support inferior to a given minimum support threshold $\vartheta$.
%For all the experiments, we keep the minimum support $\mathit{supp}_{\min}$ for the sequential features to $0.1$, thus requiring that at least $10\%$ of the traces should contain the mined sequential features.
In our experiments, the support information associated to each sequential feature is also preserved and passed as an additional information for the trace encoding phase.
Finally, for each feature $X$ and trace $\sigma$ within the log $\mathcal{L}$, the discovery algorithm returns a matrix $M$ where $M_{X,\sigma}$ denotes the relative support of $X$ within trace $\sigma$: we preserve also this information in the trace encoding step.

\paragraph*{Declarative Features}
We discover the instantiation of declarative templates via the well-known Apriori algorithm \cite{agrawalApriori}. %Using an Apriori-like approach, frequent sets of correlated activities can be efficiently discovered from an event log.
%%-- NO: A-Priori è stato il primo algoritmo ed è facile da implementare, ma non è il più efficiente. Il claim verrebbe riggettato
%Let $\Sigma$ be the set of potential activities. Let $\textbf{t} \in \Sigma^*$ be a trace over $\Sigma$, i.e., a sequence of activities executed for some process instance. An event log $\mathcal{L}$ is a multi-set over $\Sigma^*$, i.e., a trace can appear multiple times in an event log.
%The \emph{support} of a set of activities is a measure that assesses the relevance of this set in the event log.
%\begin{definition}[Support]
%The support of an activity set $A$ in an event log
%is the fraction of traces in the log that contain all the activities in $A$.
%\begin{equation*}
%\mathit{supp}(A) = \frac{\vert\mathcal{L}_{A}\vert}{\vert \mathcal{L}\vert}, \text{ where }\mathcal{L}_{A} = [\textbf{t} \in \mathcal{L} \mid \forall_{\texttt{x} \in A} \ \texttt{x} \in \textbf{t}]
%\end{equation*}
%\end{definition}
%An activity set is considered to be {\em frequent} if its support is above a given threshold $\vartheta$.
Generally speaking, let $\mathcal{A}_k$ denote the set of all frequent activity sets of size $k \in \mathbb{N}$ and let $\mathcal{C}_k$ denote the set of all {\em candidate activity sets} of size $k$ that may potentially be frequent. %The Apriori algorithm uncovers all frequent activity sets in the event log.
The algorithm starts by considering activity sets of size $k=1$ and progresses iteratively by considering activity sets of increasing sizes in each iteration. The set of candidate activity sets of size $k+1$, $\mathcal{C}_{k+1}$, is generated by joining relevant frequent activity sets from $\mathcal{A}_k$. This set can be pruned efficiently using the property that a relevant candidate activity set of size $k+1$ cannot contain an infrequent subset.
The activity sets in $\mathcal{C}_{k+1}$ that have a support above a given threshold $\vartheta$
constitute the frequent activity sets of size $k+1$ ($\mathcal{A}_{k+1}$) used in the next iteration.
We discover activity sets that have a support above $\vartheta$ in either the sub-log of deviant traces or in the sub-log of non-deviant traces.

We instantiate the former algorithm over the \declare patterns as follows:
%To discover constraints deriving from a {\declare} template with $k$ parameters, frequent activity sets of size $k$ have to be generated. Afterwards, a list of candidate constraints is generated. To do that, the {\declare} templates are instantiated by specifying
%as parameters all the possible permutations of each frequent set. %For instance,
given the
frequent activity set $\{ a, b \}$ where $a$ and $b$ are event labels, we instantiate, e.g., the \RespTmp\ template as
\RespTmp(\activity{a},\activity{b}) and \RespTmp(\activity{b},\activity{a}).
Limiting the instantiation of the candidate constraints to frequent activity sets drastically reduces the number of candidate constraints to be checked. Finally, each candidate constraint is checked separately over deviant and non-deviant traces (depending on whether it is derived from an activity set discovered from the sub-log of deviant traces or from the sub-log of non-deviant traces) to verify if it is satisfied in a percentage of traces that is above the minimum support threshold $\vartheta$. The instantiated templates resulting from the process will constitute the set of the declarative features.
%As we empirically observed that the amount of the initial patterns considerably increases for real-world datasets and is scarcer for synthetic datasets, we will vary $\vartheta$ accordingly in our experiments.

\paragraph*{Data-Aware Declarative Features} This last approach identifies data-aware declarative features, starting from the data-agnostic ones obtained as described in the previous paragraph. Starting from each data-agnostic \declare feature, we use the approach presented in \cite{DBLP:conf/bpm/LenoDM18,DBLP:journals/is/LenoDMRP20} to enrich it with a data condition as follows:

\begin{enumerate}[noitemsep]
	\item Collect the fulfilled activations of the constraint in each trace;
	\item Extract the data payload of every fulfilled activation;
	%\item Handle missing data;
	\item Encode the payloads into feature vectors in which each attribute is a feature;
 %and features of type \textit{String} are encoded into numerical features by using the one-hot encoding\footnote{One-hot encoding is a method to transform categorical data into a set of numerical features. For each unique value the categorical feature can assume, a new binary variable is created. The binary variable is encoded as 1, if the original categorical value corresponds to the one encoded by the variable, and 0, otherwise. For example, the categorical feature \textit{species} with unique values ``dog'' and ``cat'' is transformed into two features \textit{species:dog} and \textit{species:cat}. If the original observation was \textit{species = dog}, then the new feature encoding is \textit{species:dog = 1} and \textit{species:cat = 0}.};
	\item Learn a decision tree by using the feature vectors labeled as deviant or non-deviant based on whether the corresponding payloads belong to activations occurring in deviant or non-deviant traces, respectively;
	\item Create a data-aware {\declare} constraint by considering the original {\declare} constraint enriched with a data condition discriminating deviant and non-deviant traces according to the model resulting from the trained decision tree. E.g., if the initial constraint is \RespTmp(\activity{a},\activity{b}) and \activity{C} is the data condition extracted from the decision tree, the new data-aware {\declare} constraint is \RespTmp(\activity{a},\activity{b},\activity{C}). %An activity \activity{a} is an activation of the data-aware {\declare} constraint, if its payload is classified as deviant by the decision tree T.
\end{enumerate}

\subsection{Feature Selection}
\label{declare:featureselection}
The number of features generated from the previous step is, generally, too large. Therefore, it becomes important to remove the features that do not give much value for training the explanatory model. Having too many features contributes indeed to long training times, overfitting and too complex classifiers. %In addition, in this step, the selected control-flow features can be combined with data features extracted from the log.
%\subsubsection{Control-Flow Features}
The %control-flow
features discovered in the previous step are pruned via
%two different methods both using Fisher score \cite{DudaHartStork01} as basis (refer to Equation \ref{fisher}). The methods differ in the way the features are chosen. In the first method, the first $k$ features ranked according to the Fisher score are selected. In the second one (called
the coverage method, described in \cite{DBLP:books/lib/DudaHS01}. In this method, a number of features is selected by first ranking them according to the Fisher score. The Fisher score for the \textit{j-th} feature is computed as:
\begin{equation} F_j = \frac{\sum^c_{i=1}n_i{(\mu_i - \mu)}^2}{\sum^c_{i=1} n_i\sigma^2_i},
\label{fisher}
\end{equation}
where $n_i$ denotes the number of data points in class \textit{i}, $\mu_i$ and $\sigma^2_i$ denote mean and variance of class $i$ corresponding to the \textit{j-th} feature, and $\mu$ and $\sigma$ are mean and variance of all data point corresponding to the \text{j-th} feature.
Then, following the ranking, features are selected until every trace is covered by at least a fixed number of features (coverage threshold). A feature is only chosen if it covers at least one of the traces not totally covered yet.
Sequential and declarative features can be ranked and selected separately, or, in the case of the hybrid encoding, they are selected from a common ranking of sequential and declarative features.

\subsection{Trace Encoding}
\label{approach:overEnco}
In this step, each trace $\sigma$ in the input log $\mathcal{L}$ is transformed into a vector $v_\sigma$ with an associated classification label ${y}_\sigma\in\{0,1\}$, i.e., deviant (1) or non-deviant (0) trace, for training a classifier. In particular, each feature $X$ passing the feature selection phase will correspond to one distinct dimension $d_X$ within the final vector representation $v_\sigma$. %In our evaluation, we use three types of control-flow encodings (sequential, declarative and hybrid). In addition, we evaluate whether the addition of data features improves the accuracy of the classification.
%\subsubsection{Encoding the Control-Flow.}
\label{approach:traceencoding}
\paragraph*{Sequential Encoding.} For each trace $\sigma$ and given a sequential feature $X$, we set the dimension $d_X$ of the vector $v_\sigma$ to $M_{\sigma, X}$, i.e., $v_\sigma[d_X]:=M_{X,\sigma}$.

\paragraph{(Data-Aware) Declarative Encoding.} For each trace $\sigma$ and given a declarative feature $X$, we set the dimension $d_X$ of the vector $v_\sigma$ (i.e., $v_\sigma[d_X]$) to:%Each trace in the log is transformed into a numeric feature vector where each element of the vector corresponds to a {\declare} constraint (taken from the list of selected features) and has value:
\begin{itemize}[noitemsep]
\item -1, if the corresponding {\declare} constraint is violated in the trace;
\item 0, if the corresponding {\declare} constraint is vacuously satisfied in the trace;
\item $n$, if the corresponding {\declare} constraint is satisfied and activated $n$ times in the trace.
\end{itemize}
%Also in this case, the event log is transformed into a matrix of numerical values where each row corresponds to a trace and each column corresponds to a feature. For instance,
E.g., given a trace $\left\langle \activity{a},\activity{b},\activity{c},\activity{a},\activity{b},\activity{c},\activity{d},\activity{a},\activity{b}\right\rangle$:
\begin{itemize}[noitemsep]
\item constraint \RespTmp(a,c) is violated, since the third activation leads to a violation and is encoded as -1;
\item constraint \RespTmp(a,b) is satisfied and activated 3 times and is encoded as 3;
\item constraint \RespTmp(e,b) is vacuously satisfied and is encoded as 0.
\end{itemize}
We also adopt the same approach for representing data-aware declarative features. E.g., given a trace $\left\langle \activity{a} \textit{\{color = white\}}, \activity{c}, \activity{b} \textit{\{color = black\}}, \activity{c}, \activity{d}, \activity{a} \textit{\{color = white\}}, \activity{c} \right\rangle$, we have:
%and a data condition \activity{C} \textit{(color = white) $\Rightarrow$ 1} leading to a positive (i.e., deviant) label ($y_\sigma=1$), then:
\begin{itemize}[noitemsep]
	\item constraint \RespTmp(a,c,color = white) is satisfied and activated twice, and is hence encoded as 2;
	\item constraint \RespTmp(a,d,color = white) is violated, since the second occurrence of \activity{a} is not eventually followed by \activity{d}, and is encoded as -1;
	\item constraint \RespTmp(b,c,color = white) is vacuously satisfied and encoded as 0 (\activity{b} is not an activation, because the data condition \textit{color = white} does not hold on its payload).
\end{itemize}

%\paragraph{Alternative declarative encoding scheme.} In addition to the encoding scheme above, one of the the alternative encoding schemes is the following:

%\begin{itemize}[noitemsep]
%\item a feature has value -1, if the corresponding {\declare} constraint is violated in the trace;
%\item a feature has value 0, if the corresponding {\declare} constraint is vacuously satisfied in the trace;
%\item a feature has value 1, if the corresponding {\declare} constraint is satisfied and activated once in a trace;
%\item a feature has value 2, if the corresponding {\declare} constraint is satisfied and activated multiple times in the trace.
%\end{itemize}

%The difference between experimented and alternative encoding scheme is in the fact that every constraint with activation count of 2 or larger is encoded as 2. Using this scheme, the encodings can be looked at as classes instead of numerical values, due to having a maximum of four different encodings. This scheme, however loses possible information gained from having more than two activations.

\paragraph*{Hybrid Encoding} Each trace is encoded into a numerical feature vector as explained in the previous two paragraphs depending on whether the feature is sequential or declarative.

In our experiments, all the encodings explained so far have been considered with and without the data features introduced defined and selected in the previous steps of the pipeline.

\subsection{Model Training}
For each type of encoding described, a classifier is trained to both classify new unseen traces (thus being able to evaluate the performance of the classification) and explain the classification with explicit rules. For this reason, the chosen classifiers are \emph{white-box} classifiers.

\subsection{Rule Extraction}
\label{exp:re}
%Ripper$k$ inherently provides a set of decision rules as output.
As output of a business process deviance mining approach, it is important to provide a set of rules describing the differences between deviant and non-deviant traces. To identify these rules, we use the white-box classifiers Ripper$k$ (Repeated Incremental Pruning to Produce Error Reduction) \cite{Cohen95fasteffective} and decision trees \cite{Quinlan:1993:CPM:152181}. Such models directly provide the classification rules: while Ripper directly provides the rules in output, they are reconstructed from a decision tree by considering the conjunction of the atomic conditions encountered on each path from the root of the tree to the leaves labeled as deviant (each path is a separate rule). Conditions on the same feature are simplified by merging them (when possible) and by removing subsumed conditions.

\subsection{The Presented Pipeline in the Context of the Motivating Example}
\label{ssec:application}
Consider the motivating example presented in Section~\ref{sec:problem}. By applying the business process deviance mining pipeline with sequential features, analysts would get as outcome that deviant traces are those for which the pattern $\langle\activity{ER}, \activity{RCR}, \activity{NC} \rangle$ is repeated more than twice (i.e., it is a tandem repeat with frequency greater than 2). However, although many of the deviant traces seem to be captured by this explanation, it could not be sufficient for characterizing all of them.
% (the classification accuracy is not $100\%$).

The analysts can then try to apply the approach with declarative features. The outcome, in this case, is that deviant traces are those for which activity $\activity{Resolve Simple Requests}$ is eventually followed by $\activity{Document Request}$ (i.e., for which \RespTmp(\activity{RSR}, \activity{DR}) is satisfied). Still, this explanation is not sufficient for capturing all the deviant traces.

By applying the approach with hybrid features, the analysts are finally able to get an explanation using both sequential and declarative patterns, and covering almost all traces, i.e., there are two situations in which the process executions take longer:
\begin{itemize}[noitemsep]
\item when, in the case of complex requests, more than two iterations are carried out with the customer, until the customer is satisfied.
\item when customer support operators procrastinate writing reports related to simple request resolutions. Indeed, differently from complex requests, for which operators tend to complete the documentation immediately after the resolution, reports for simple requests tend to be delayed and the requests cannot be closed quickly.
\end{itemize}

Since the control-flow patterns are not able to completely discriminate deviant from non-deviant traces, analysts could decide to add data features. Using these additional features, they find that a first cause of deviance is the same as the one they found without considering the data perspective, i.e., there is a deviance when pattern  $\langle\activity{ER}, \activity{RCR}, \activity{NC} \rangle$ is repeated more than twice. However, with the addition of \textit{pure data} features, it would be possible to see that requests received in July and August took more time to be processed. Also, by adding features based on data-aware {\declare} constraints, analysts could find explanations telling that constraint \RespTmp(\activity{RSR},\activity{DR}, \textit{resource = D}) characterizes a deviant trace better than the initial {\declare} constraint without the data condition. This means that, if \activity{RSR} was performed by a specific resource (D), the execution time was longer.

The addition of features based on data attributes help analysts to enrich and refine the deviance causes previously identified. In this case, analysts were able to identify three situations where process executions take considerably longer:
\begin{itemize}[noitemsep]
\item when the request is received during the summer;
\item when a customer support operator (resource D) procrastinate writing reports related to simple request resolutions;
\item when, in the case of complex requests, more than two iterations are carried out with the customer.
\end{itemize}

%\subsection{Extraction of Results of Deviance Mining}
%\label{subsec:extraction}

\section{Evaluation}
\label{sec:allevaluations}
In this section, we describe the experimentation we have carried out to evaluate our pipeline for exploring business process deviance. The source code is available on GitHub at \url{https://github.com/jackbergus/CompleteDevianceMining}. %The full set of results can be found within the same repository in \textit{ExperimentResults} folder, where also included are results of measuring the metrics on training data and standard deviations for each of the metrics.
For the construction of the decision trees used in the experiments, we used the Python library \textit{scikit-learn} \cite{scikit-learn}. For Ripper$k$, we used \textit{JRip} available in Weka \cite{hall09:_weka_data_minin_softw}. %\footnote{In all the experiments, the max depth of the tree was fixed to 10 and the leaves of the decision tree were forced to be of at least size 5.}
%\footnote{The number of optimization steps is set to 2.}

\subsection{Research Questions}\label{sssec:rq}
%\textbf{\color{gray}Evaluating the Impact of Control-Flow Features on Business Process Deviance Mining}
%\label{eval:first}
%
%{\color{red}[Rephrase the research questions]}
%
In order to investigate the impact of different types of encodings (sequential, declarative and hybrid with and without data) and of the chosen classifier (Ripper$k$ and decision trees) on their ability to accurately discriminate between non-deviant and deviant executions of a process, the following research questions were considered:

\begin{itemize}[noitemsep]
\item [\textbf{RQ1.}] How does the choice of the trace encoding affect the accuracy of the classification of traces into deviant and non-deviant, in the proposed pipeline?
\item [\textbf{RQ2.}] How does the choice of the classifier affect the accuracy of the classification of traces into deviant and non-deviant, in the proposed pipeline?
\item [\textbf{RQ3.}] How does the choice of the trace encoding affect the conciseness of the rules describing the differences between deviant and non-deviant traces?
\item [\textbf{RQ4.}] How does the choice of the classifier affect the conciseness of the rules describing the differences between deviant and non-deviant traces?
\end{itemize}
\textbf{RQ1} and \textbf{RQ2} evaluate how different types of features and different classifiers affect the accuracy of the deviance mining task. \textbf{RQ3} and \textbf{RQ4}, instead, focus on how different types of features and different classifiers affect the conciseness of the rules used to explain deviances. This is very relevant from the user point of view, since more concise rules are more understandable and might generalize better over noisy data \cite{DBLP:journals/ai/Miller19}.

\subsection{Datasets and Dataset Labelings}\label{sssec:datasets}
The evaluation has been carried out using 9 real-life logs. In particular, we considered several BPI Challenge Datasets (one log from 2011, one log from 2012, five logs from 2015), the Sepsis Case event log~\cite{sepsislog}, and the Traffic Fine Management Process log~\cite{traffic}.

The Sepsis Cases event log (\texttt{sepsis}) collects cases of patients with symptoms of sepsis from a Dutch hospital.
%As no trace attributes were available, we exploited event attributes for the labelings. %The Sepsis Cases log was labeled based on two different criteria: one based on procedural patterns ($\textit{Sepsis}_{Proc}$) and one based on declarative patterns ($\textit{Sepsis}_{Decl}$):
%and a third one based on temporal aspects ($\textit{Sepsis}_{ER}$):
Log traces were labeled as:
\begin{itemize}[noitemsep]
	\item \texttt{decl}: \RespTmp(\texttt{IV Antibiotics}, \texttt{Leucocytes}), \RespTmp(\texttt{LacticAcid}, \texttt{IV Antibiotics}) and \RespTmp(\texttt{ER Triage}, \texttt{CRP}) are satisfied non-vacuously.
	\item \texttt{mr\_tr}: sequence ``\texttt{Admission NC}, \texttt{CRP}, \texttt{Leucocytes}'' occurs at least once within the trace.
	\item \texttt{mra\_tra}: ``\texttt{IV Liquid}, \texttt{LacticAcid}, \texttt{Leucocytes}'' occur in any order twice within the trace, with interleaving.
	\item \texttt{payload}: Trace attribute \texttt{DisfuncOrg}=\textit{True}.
	%\item \texttt{proc}: ``\texttt{Admission NC}, \texttt{Leucocytes}'' is an exact subsequence of the trace.
	%\item the patient returned to the emergency room within 28 days after having been released ($\textit{Sepsis}_{ER}$).
\end{itemize}

The Traffic Fine Management Process log (\texttt{traffic}) describes some events related to the Italian fine system for road traffic. Log traces were labeled as:
\begin{itemize}
	\item \texttt{decl}: \RespTmp(\texttt{Insert Date Appeal to Prefecture}, \texttt{Add penalty}) is satisfied non-vacuously.
	\item \texttt{mr\_tr}: sequence ``\texttt{Add penalty} \texttt{Payment}'' occurs at least once within the trace.
	\item \texttt{mra\_tra}: ``\texttt{Create Fine}, \texttt{Payment}'' occur in any order twice within the trace, with interleaving.
	\item \texttt{p\_Art157}: Trace attribute \texttt{article}=\textit{157}.
	\item \texttt{p\_Pay36}: Event attribute \texttt{paymentAmount}=\textit{36} at least once.
%	\item \texttt{proc}:
%	``\texttt{Create Fine}, \texttt{Payment}'' is an exact subsequence of the trace.
\end{itemize}

The BPI Challenge 2011 log (\texttt{bpi11}) contains data about a Dutch academic hospital. Each trace represents the clinical history of a patient. Log traces were labeled as:
\begin{itemize}
	\item \texttt{decl}: \InitTmp(\texttt{Outpatient follow-up consultation}) is satisfied.
	\item \texttt{mr\_tr}: sequence ``\texttt{SGOT}, \texttt{SGPT}, \texttt{Milk acid dehydrogenase LDH}, \texttt{Leukocytes electronic count}'' occurs at least once within the trace.
	\item \texttt{mra\_tra}: ``\texttt{Assumption Laboratory}, \texttt{Milk acid dehydrogenase LDH}'' occur in any order twice within the trace, with interleaving.
	\item \texttt{p\_M13} \cite{Nguyen2014}: Trace attribute \textsf{Diagnosis code}=\textit{M13}.
	\item \texttt{p\_T101} \cite{Nguyen2014}: Trace attribute \textsf{Treatment code}=\textit{101}.
%	\item \texttt{proc}: ``\texttt{Un-conjugated Bilirubin} \texttt{Bilirubin:Total}, \texttt{Glucose}'' is an exact subsequence of the trace.
\end{itemize}

The BPI Challenge 2012 log (\texttt{bpi12}) contains data about a Dutch Financial Institute. The process represented in the event log is an application process for a personal loan or overdraft within a global financing organization. Log traces were labeled as:
\begin{itemize}
	\item \texttt{decl}: \PrecTmp(\texttt{O\_ACCEPTED}, \texttt{A\_APPROVED}) is satisfied non-vacuously.
	\item \texttt{mr\_tr}: sequence ``\texttt{O\_SENT}, \texttt{Request Completed}, \texttt{Recall incomplete dossiers}'' occurs at least once within the trace.
	\item \texttt{mra\_tra}: ``\texttt{Handling leads}, \texttt{Request Completed}'' occur in any order three times within the trace, with interleaving.
	\item \texttt{p\_45000}: Trace attribute \texttt{AMOUNT\_REQ}=\textit{45000}.
	\item \texttt{p\_6500}: Trace attribute \texttt{AMOUNT\_REQ}=\textit{6500}.
%	\item \texttt{proc}:
%	``\texttt{Request Completed}, \texttt{Handling leads}'' is an exact subsequence of the trace.
\end{itemize}

The BPI Challenge 2015 logs (\texttt{bpi15A}-\texttt{bpi15E}) come from five distinct Dutch municipalities. The logs pertain to the assessment of building permit applications in each municipality. Each log was labeled as:
\begin{itemize}
		\item \texttt{decl}: \ExiTmp(\texttt{01\_HOOFD\_011}) is satisfied.
	\item \texttt{mr\_tr}: sequence ``\texttt{08\_AWB45\_005}, \texttt{01\_HOOFD\_200}'' occurs at least once within the trace.
	\item \texttt{mra\_tra}: ``\texttt{01\_HOOFD\_030\_1}, \texttt{01\_HOOFD\_510\_1}'' occur in any order twice within the trace, with interleaving.
		\item \texttt{payload}: Trace attribute \texttt{monitoringResource}=$x$, where $x$ depends on the resource monitored within the dataset (\texttt{A:}\textit{560925}, \texttt{B:}\textit{4634935}, \texttt{C:}\textit{3442724}, \texttt{D:}\textit{560812}, \texttt{E:}\textit{560608}).
%	\item \texttt{proc}:
%	``\texttt{01\_HOOFD\_010}, \texttt{01\_HOOFD\_015}'' is an exact subsequence of the trace.
	%\item \textbf{Declarative}$_3$: \RespTmp(\texttt{01\_HOOFD\_011}, \texttt{02\_DRZ\_010}) satisfied non-vacuously.
\end{itemize}

\subsection{Results}
\label{sssec:procedure}

%\begin{figure}[t!]
%\includegraphics[width=1\textwidth]{figures/crossvalidpipe}
%\centering
%\caption{Cross-validation pipeline}
%\label{fig:cross}
%\end{figure}

To answer our research questions, we run the proposed pipeline using different configurations. In particular, we use as classifiers Ripper$k$ and decision trees, and  3-fold cross-validation to train and test them. In addition, we use grid search for hyper-parameter tuning. We used all the feature encodings introduced in Section~\ref{sec:allapproaches}. In the feature discovery step, we set $\vartheta=0.3$ as minimum support threshold to generate a sufficiently high number of features.

To answer research questions \textbf{RQ1} and \textbf{RQ2}, we use standard measures, i.e., \emph{precision}, \emph{recall}, \emph{F1}, and \emph{AUC}, to estimate the accuracy of the different encodings and classifiers. In particular, we evaluate these metrics, for each classifier, each encoding (Individual Activities (IA), TR, TRA, MR, MRA, Declare, Hybrid (H), (Pure) Data, Data + Individual Activities (IA), Data+TR, Data+TRA, Data+MR, Data+MRA, Data+Declare, Data+Hybrid (H), Data+ Data-aware Declare (DeclD), H+DeclD, H+Data+DeclD), each dataset and each labeling, and we average them over the different datasets.

Tables \ref{tab:RDRipperk} and \ref{tab:decision tree} provide the results obtained using as classifiers Ripper$k$ and decision trees, respectively. The results show that logs labeled using declarative rules provide, in general, the best performances with the declarative encodings. In some cases, MRA patterns also are able to predict well the declarative labelings. This is reasonable since MRA patterns sit in the middle between procedural and declarative patterns and are capable to characterize ``unstructured'' behavior. On the other hand, in most of the cases, MR and TR patterns guarantee a good performance for procedural labelings. The \texttt{mra\_tra} labelings tested can be, instead, accurately predicted using very little information (individual activities are sufficient) and the accuracy measures are equal to 1 for almost all the encodings. Notice that for both declarative and procedural labelings, using the hybrid encoding, it is possible to predict the correct label in all cases when using Ripper$k$ and with a very high accuracy also when using decision trees.

As expected, the data features are less accurate in predicting control-flow labels and work better than control-flow features when predicting labels based on data attributes. However, even in the latter case, the addition of some control-flow information improves the accuracy of the results. The tested classifiers have comparable performance in terms of accuracy. However, Ripper$k$ seems to better exploit the availability of hybrid features with respect to decision trees.

\begin{table}[t!]
	\caption{Average accuracy metrics using Ripper$k$}\label{tab:RDRipperk}
	{\setlength\tabcolsep{1.85pt}\tiny\begin{tabular}{l|llll|llll|llll|llll|llll}
			\toprule
			\multirow{2}{*}{Labelling} & \multicolumn{4}{c}{IA} & \multicolumn{4}{c}{MR} & \multicolumn{4}{c}{TR} & \multicolumn{4}{c}{MRA} & \multicolumn{4}{c}{TRA} \\
			{}  & Prec & Rec & F1 & AUC  & Prec & Rec & F1 & AUC  & Prec & Rec & F1 & AUC  & Prec & Rec & F1 & AUC  & Prec & Rec & F1 & AUC \\
			\midrule
			\texttt{decl} & 0.96 & 0.99 & 0.99 & 0.97 & 0.98 & 0.98 & 0.98 & 0.98 & 0.95 & 0.99 & 0.99 & 0.97 & 0.99 & 0.98 & 0.99 & 0.98 & 0.95 & 0.98 & 0.98 & 0.96\\
			\texttt{mr\_tr} & 0.88 & 0.89 & 0.89 & 0.89 & \textbf{1.00}  & \textbf{1.00}  & \textbf{1.00}  & \textbf{1.00}  & \textbf{1.00}  & \textbf{1.00}  & \textbf{1.00}  & \textbf{1.00}  & 0.99 & 0.97 & 0.98 & 0.98 & 0.98 & 0.98 & 0.99 & 0.98\\
			\texttt{mra\_tra} & \textbf{1.00}  & \textbf{1.00}  & \textbf{1.00}  & \textbf{1.00}  & \textbf{1.00}  & \textbf{1.00}  & \textbf{1.00}  & \textbf{1.00}  & \textbf{1.00}  & \textbf{1.00}  & \textbf{1.00}  & \textbf{1.00}  & \textbf{1.00}  & \textbf{1.00}  & \textbf{1.00}  & \textbf{1.00}  & \textbf{1.00}  & \textbf{1.00}  & \textbf{1.00}  & \textbf{1.00} \\
			\texttt{payload} & 0.34 & {\color{red} 0.21}  & {\color{red} 0.48}  & 0.26 & 0.40 & 0.29 & 0.59 & 0.32 & 0.32 & 0.23 & 0.52 & 0.25 & 0.47 & 0.30 & 0.63 & 0.33 & {\color{red} 0.31}  & {\color{red} 0.21}  & {\color{red} 0.48}  & {\color{red}0.23}\\
%			\texttt{proc} & 0.78 & 0.67 & 0.82 & 0.70 & \textbf{1.00}  & \textbf{1.00}  & \textbf{1.00}  & \textbf{1.00}  & 0.81 & 0.70 & 0.85 & 0.72 & 0.98 & 0.98 & 0.99 & 0.98 & 0.79 & 0.70 & 0.85 & 0.72\\
			\bottomrule
	\end{tabular}}

	{\setlength\tabcolsep{4pt} \tiny\begin{tabular}{l|llll|llll|llll|llll}
			\toprule
			\multirow{2}{*}{Labelling} & \multicolumn{4}{c}{Declare} & \multicolumn{4}{c}{H} & \multicolumn{4}{c}{Data} & \multicolumn{4}{c}{DeclD} \\
			{}  & Prec & Rec & F1 & AUC  & Prec & Rec & F1 & AUC  & Prec & Rec & F1 & AUC  & Prec & Rec & F1 & AUC \\
			\midrule
			\texttt{decl} & \textbf{1.00}  & \textbf{1.00}  & \textbf{1.00}  & \textbf{1.00}  & \textbf{1.00}  & \textbf{1.00}  & \textbf{1.00}  & \textbf{1.00}  & {\color{red} 0.81}  & {\color{red} 0.82}  & {\color{red} 0.80}  & {\color{red} 0.81} & \textbf{1.00}  & \textbf{1.00}  & \textbf{1.00}  & \textbf{1.00} \\
			\texttt{mr\_tr} & 0.99 & \textbf{1.00}  & 0.99 & 0.99 & \textbf{1.00}  & \textbf{1.00}  & \textbf{1.00}  & \textbf{1.00}  & {\color{red} 0.76}  & {\color{red} 0.74}  & {\color{red} 0.77}  & {\color{red} 0.77}  & 0.98 & \textbf{1.00}  & \textbf{1.00}  & 0.99\\
			\texttt{mra\_tra} & \textbf{1.00}  & \textbf{1.00}  & \textbf{1.00}  & \textbf{1.00}  & \textbf{1.00}  & \textbf{1.00}  & \textbf{1.00}  & \textbf{1.00}  & {\color{red} 0.84}  & {\color{red} 0.83}  & {\color{red} 0.86}  & {\color{red} 0.82} & \textbf{1.00}  & \textbf{1.00}  & \textbf{1.00}  & \textbf{1.00} \\
			\texttt{payload} & 0.45 & 0.33 & 0.57 & 0.36 & 0.35 & 0.49 & 0.67 & 0.37 & 0.83 & 0.78 & 0.88 & 0.81 & 0.74 & 0.86 & 0.90 & 0.77\\
%			\texttt{proc} & \textbf{1.00}  & \textbf{1.00}  & \textbf{1.00}  & \textbf{1.00}  & \textbf{1.00}  & \textbf{1.00}  & \textbf{1.00}  & \textbf{1.00}  & {\color{red}0.55} & {\color{red} 0.43}  & {\color{red} 0.63}  & {\color{red} 0.48} & \textbf{1.00}  & \textbf{1.00}  & \textbf{1.00}  & \textbf{1.00} \\
			\bottomrule
	\end{tabular}}

	{\setlength\tabcolsep{1.85pt}\tiny\begin{tabular}{l|llll|llll|llll|llll|llll}
			\toprule
			\multirow{2}{*}{Labelling} & \multicolumn{4}{c}{Data+IA} & \multicolumn{4}{c}{Data+MR} & \multicolumn{4}{c}{Data+TR} & \multicolumn{4}{c}{Data+MRA} & \multicolumn{4}{c}{Data+TRA} \\
			{}  & Prec & Rec & F1 & AUC  & Prec & Rec & F1 & AUC  & Prec & Rec & F1 & AUC  & Prec & Rec & F1 & AUC  & Prec & Rec & F1 & AUC \\
			\midrule
			\texttt{decl} & 0.95 & 0.98 & 0.98 & 0.96 & 0.99 & 0.98 & 0.99 & 0.98 & 0.95 & 0.98 & 0.98 & 0.96 & 0.99 & 0.97 & 0.98 & 0.98 & 0.95 & 0.99 & 0.98 & 0.96\\
			\texttt{mr\_tr} & 0.94 & 0.87 & 0.93 & 0.88 & 0.99 & \textbf{1.00}  & \textbf{1.00}  & 0.99 & \textbf{1.00}  & 0.95 & 0.98 & 0.96 & 0.99 & 0.97 & 0.99 & 0.98 & 0.98 & 0.94 & 0.98 & 0.94\\
			\texttt{mra\_tra} & \textbf{1.00}  & \textbf{1.00}  & \textbf{1.00}  & \textbf{1.00}  & \textbf{1.00}  & \textbf{1.00}  & \textbf{1.00}  & \textbf{1.00}  & \textbf{1.00}  & \textbf{1.00}  & \textbf{1.00}  & \textbf{1.00}  & \textbf{1.00}  & \textbf{1.00}  & \textbf{1.00}  & \textbf{1.00}  & \textbf{1.00}  & \textbf{1.00}  & \textbf{1.00}  & \textbf{1.00} \\
			\texttt{payload} & \textbf{0.88}  & 0.83 & \textbf{0.91}  & \textbf{0.84} & 0.84 & 0.80 & 0.90 & 0.83 & 0.84 & 0.77 & 0.88 & 0.81 & 0.83 & 0.75 & 0.87 & 0.77 & 0.87 & 0.79 & 0.89 & \textbf{0.84}\\
%			\texttt{proc} & 0.71 & 0.71 & 0.83 & 0.68 & 0.95 & 0.95 & 0.98 & 0.95 & 0.82 & 0.66 & 0.83 & 0.68 & 0.98 & 0.95 & 0.98 & 0.96 & 0.82 & 0.73 & 0.84 & 0.75\\
			\bottomrule
	\end{tabular}}

	{\setlength\tabcolsep{1.85pt}\tiny\begin{tabular}{l|llll|llll|llll|llll|llll}
			\toprule
			\multirow{2}{*}{Labelling} & \multicolumn{4}{c}{Data+Declare} & \multicolumn{4}{c}{Data+DeclD} & \multicolumn{4}{c}{Data+H} & \multicolumn{4}{c}{H+DeclD} & \multicolumn{4}{c}{H+Data+DeclD} \\
			{}  & Prec & Rec & F1 & AUC  & Prec & Rec & F1 & AUC  & Prec & Rec & F1 & AUC  & Prec & Rec & F1 & AUC  & Prec & Rec & F1 & AUC \\
			\midrule
			\texttt{decl} & \textbf{1.00}  & \textbf{1.00}  & \textbf{1.00}  & \textbf{1.00}  & \textbf{1.00}  & \textbf{1.00}  & \textbf{1.00}  & \textbf{1.00}  & \textbf{1.00}  & \textbf{1.00}  & \textbf{1.00}  & \textbf{1.00}  & \textbf{1.00}  & \textbf{1.00}  & \textbf{1.00}  & \textbf{1.00}  & \textbf{1.00}  & \textbf{1.00}  & \textbf{1.00}  & \textbf{1.00} \\
			\texttt{mr\_tr} & 0.98 & 0.99 & 0.99 & 0.99 & 0.98 & \textbf{1.00}  & \textbf{1.00}  & 0.99 & \textbf{1.00}  & \textbf{1.00}  & \textbf{1.00}  & \textbf{1.00}  & \textbf{1.00}  & \textbf{1.00}  & \textbf{1.00}  & \textbf{1.00}  & \textbf{1.00}  & \textbf{1.00}  & \textbf{1.00}  & \textbf{1.00} \\
			\texttt{mra\_tra} & \textbf{1.00}  & \textbf{1.00}  & \textbf{1.00}  & \textbf{1.00}  & \textbf{1.00}  & \textbf{1.00}  & \textbf{1.00}  & \textbf{1.00}  & \textbf{1.00}  & \textbf{1.00}  & \textbf{1.00}  & \textbf{1.00}  & \textbf{1.00}  & \textbf{1.00}  & \textbf{1.00}  & \textbf{1.00}  & \textbf{1.00}  & \textbf{1.00}  & \textbf{1.00}  & \textbf{1.00} \\
			\texttt{payload} & 0.73 & 0.85 & 0.90 & 0.76 & 0.80 & 0.81 & 0.89 & 0.77 & 0.72 & \textbf{0.88}  & \textbf{0.91}  & 0.77 & 0.75 & 0.82 & 0.89 & 0.76 & 0.76 & 0.82 & 0.89 & 0.76\\
%			\texttt{proc} & \textbf{1.00}  & \textbf{1.00}  & \textbf{1.00}  & \textbf{1.00}  & \textbf{1.00}  & \textbf{1.00}  & \textbf{1.00}  & \textbf{1.00}  & \textbf{1.00}  & \textbf{1.00}  & \textbf{1.00}  & \textbf{1.00}  & \textbf{1.00}  & \textbf{1.00}  & \textbf{1.00}  & \textbf{1.00}  & \textbf{1.00}  & \textbf{1.00}  & \textbf{1.00}  & \textbf{1.00} \\
			\bottomrule
	\end{tabular}}
	
\end{table}

\begin{table}[t!]
	\caption{Average accuracy metrics using decision tree}\label{tab:decision tree}
	{\setlength\tabcolsep{1.85pt} \tiny\begin{tabular}{l|llll|llll|llll|llll|llll}
			\toprule
			\multirow{2}{*}{Labelling} & \multicolumn{4}{c}{IA} & \multicolumn{4}{c}{MR} & \multicolumn{4}{c}{TR} & \multicolumn{4}{c}{MRA} & \multicolumn{4}{c}{TRA} \\
			{}  & Prec & Rec & F1 & AUC  & Prec & Rec & F1 & AUC  & Prec & Rec & F1 & AUC  & Prec & Rec & F1 & AUC  & Prec & Rec & F1 & AUC \\
			\midrule
			\texttt{decl} & 0.95 & 0.97 & 0.98 & 0.96 & 0.98 & 0.98 & \textbf{0.99}  & 0.98 & 0.95 & 0.96 & 0.98 & 0.95 & \textbf{0.99}  & 0.98 & \textbf{0.99}  & 0.98 & 0.95 & 0.97 & \textbf{0.99}  & 0.96\\
			\texttt{mr\_tr} & 0.94 & 0.93 & 0.97 & 0.94 & \textbf{1.00}  & \textbf{1.00}  & \textbf{1.00}  & \textbf{1.00}  & \textbf{1.00}  & \textbf{1.00}  & \textbf{1.00}  & \textbf{1.00}  & 0.99 & 0.97 & \textbf{1.00}  & 0.98 & 0.99 & 0.98 & 0.99 & 0.99\\
			\texttt{mra\_tra} & \textbf{1.00}  & \textbf{1.00}  & \textbf{1.00}  & \textbf{1.00}  & \textbf{1.00}  & \textbf{1.00}  & \textbf{1.00}  & \textbf{1.00}  & \textbf{1.00}  & \textbf{1.00}  & \textbf{1.00}  & \textbf{1.00}  & \textbf{1.00}  & \textbf{1.00}  & \textbf{1.00}  & \textbf{1.00}  & \textbf{1.00}  & \textbf{1.00}  & \textbf{1.00}  & \textbf{1.00} \\
			\texttt{payload} & 0.34 & 0.28 & 0.67 & 0.29 & 0.41 & 0.33 & 0.70 & 0.34 & {\color{red} 0.33}  & {\color{red} 0.26}  & 0.67 & {\color{red}0.28} & 0.44 & 0.35 & 0.70 & 0.37 & {\color{red} 0.33}  & 0.28 & {\color{red} 0.66}  & {\color{red}0.28}\\
%			\texttt{proc} & 0.70 & 0.64 & 0.83 & 0.66 & \textbf{0.99}  & 0.95 & \textbf{0.99}  & \textbf{0.97} & 0.77 & 0.68 & 0.86 & 0.71 & 0.98 & 0.93 & 0.98 & 0.95 & 0.74 & 0.68 & 0.85 & 0.70\\
			\bottomrule
	\end{tabular}}

	{\setlength\tabcolsep{4pt} \tiny\begin{tabular}{l|llll|llll|llll|llll}
			\toprule
			\multirow{2}{*}{Labelling} & \multicolumn{4}{c}{Declare} & \multicolumn{4}{c}{H} & \multicolumn{4}{c}{Data} & \multicolumn{4}{c}{DeclD} \\
			{}  & Prec & Rec & F1 & AUC  & Prec & Rec & F1 & AUC  & Prec & Rec & F1 & AUC  & Prec & Rec & F1 & AUC \\
			\midrule
			\texttt{decl} & 0.98 & \textbf{0.99}  & \textbf{0.99}  & 0.98 & 0.98 & \textbf{0.99}  & \textbf{0.99}  & 0.98 & {\color{red} 0.82}  & {\color{red} 0.79}  & {\color{red} 0.82}  & {\color{red}0.79} & 0.97 & 0.96 & 0.98 & 0.97\\
			\texttt{mr\_tr} & 0.93 & \textbf{1.00}  & 0.95 & 0.96 & \textbf{1.00}  & \textbf{1.00}  & \textbf{1.00}  & \textbf{1.00}  & {\color{red} 0.80}  & {\color{red} 0.79}  & {\color{red} 0.87}  & {\color{red}0.79} & 0.99 & 0.99 & 0.99 & 0.99\\
			\texttt{mra\_tra} & \textbf{1.00}  & \textbf{1.00}  & \textbf{1.00}  & \textbf{1.00}  & 0.99 & \textbf{1.00}  & 0.98 & 0.99 & {\color{red} 0.82}  & {\color{red} 0.77}  & {\color{red} 0.86}  & {\color{red}0.79} & \textbf{1.00}  & \textbf{1.00}  & \textbf{1.00}  & \textbf{1.00} \\
			\texttt{payload} & 0.42 & 0.51 & 0.79 & 0.40 & 0.42 & 0.52 & 0.78 & 0.41 & 0.84 & 0.79 & 0.90 & 0.81 & 0.68 & 0.84 & 0.90 & 0.72\\
%			\texttt{proc} & 0.76 & \textbf{1.00}  & 0.97 & 0.84 & 0.73 & \textbf{1.00}  & 0.97 & 0.82 & {\color{red} 0.54}  & {\color{red} 0.41}  & {\color{red} 0.75}  & {\color{red}0.45} & 0.76 & \textbf{1.00}  & 0.97 & 0.84\\
			\bottomrule
	\end{tabular}}

	{\setlength\tabcolsep{1.85pt} \tiny\begin{tabular}{l|llll|llll|llll|llll|llll}
			\toprule
			\multirow{2}{*}{Labelling} & \multicolumn{4}{c}{Data+IA} & \multicolumn{4}{c}{Data+MR} & \multicolumn{4}{c}{Data+TR} & \multicolumn{4}{c}{Data+MRA} & \multicolumn{4}{c}{Data+TRA} \\
			{}  & Prec & Rec & F1 & AUC  & Prec & Rec & F1 & AUC  & Prec & Rec & F1 & AUC  & Prec & Rec & F1 & AUC  & Prec & Rec & F1 & AUC \\
			\midrule
			\texttt{decl} & 0.95 & 0.98 & 0.98 & 0.96 & 0.98 & 0.97 & \textbf{0.99}  & 0.98 & 0.95 & 0.97 & 0.98 & 0.96 & 0.98 & 0.98 & \textbf{0.99}  & \textbf{0.99}  & 0.96 & 0.97 & \textbf{0.99}  & 0.96\\
			\texttt{mr\_tr} & 0.94 & 0.94 & 0.96 & 0.94 & \textbf{1.00}  & \textbf{1.00}  & \textbf{1.00}  & \textbf{1.00}  & \textbf{1.00}  & \textbf{1.00}  & \textbf{1.00}  & \textbf{1.00}  & 0.99 & 0.97 & 0.99 & 0.98 & 0.99 & 0.98 & \textbf{1.00}  & 0.99\\
			\texttt{mra\_tra} & \textbf{1.00}  & \textbf{1.00}  & \textbf{1.00}  & \textbf{1.00}  & \textbf{1.00}  & \textbf{1.00}  & \textbf{1.00}  & \textbf{1.00}  & \textbf{1.00}  & \textbf{1.00}  & \textbf{1.00}  & \textbf{1.00}  & \textbf{1.00}  & \textbf{1.00}  & \textbf{1.00}  & \textbf{1.00}  & \textbf{1.00}  & \textbf{1.00}  & \textbf{1.00}  & \textbf{1.00} \\
			\texttt{payload} & \textbf{0.86}  & 0.84 & \textbf{0.91}  & \textbf{0.84} & 0.84 & 0.83 & \textbf{0.91}  & 0.83 & 0.84 & 0.83 & \textbf{0.91}  & 0.83 & 0.83 & 0.81 & 0.90 & 0.82 & 0.84 & 0.83 & \textbf{0.91}  & 0.83\\
%			\texttt{proc} & 0.70 & 0.64 & 0.84 & 0.66 & \textbf{0.99}  & 0.94 & \textbf{0.99}  & 0.96 & 0.80 & 0.68 & 0.89 & 0.72 & \textbf{0.99}  & 0.95 & 0.98 & 0.96 & 0.80 & 0.69 & 0.89 & 0.72\\
			\bottomrule
	\end{tabular}}

	{\setlength\tabcolsep{1.85pt} \tiny\begin{tabular}{l|llll|llll|llll|llll|llll}
			\toprule
			\multirow{2}{*}{Labelling} & \multicolumn{4}{c}{Data+Declare} & \multicolumn{4}{c}{Data+DeclD} & \multicolumn{4}{c}{Data+H} & \multicolumn{4}{c}{H+DeclD} & \multicolumn{4}{c}{H+Data+DeclD} \\
			{}  & Prec & Rec & F1 & AUC  & Prec & Rec & F1 & AUC  & Prec & Rec & F1 & AUC  & Prec & Rec & F1 & AUC  & Prec & Rec & F1 & AUC \\
			\midrule
			\texttt{decl} & 0.98 & \textbf{0.99}  & \textbf{0.99}  & 0.98 & 0.97 & 0.96 & 0.98 & 0.97 & 0.98 & \textbf{0.99}  & \textbf{0.99}  & 0.98 & 0.97 & 0.96 & 0.98 & 0.97 & 0.97 & 0.96 & 0.98 & 0.96\\
			\texttt{mr\_tr} & 0.98 & \textbf{1.00}  & \textbf{1.00}  & 0.99 & 0.98 & 0.98 & 0.99 & 0.99 & \textbf{1.00}  & \textbf{1.00}  & \textbf{1.00}  & \textbf{1.00}  & \textbf{1.00}  & \textbf{1.00}  & \textbf{1.00}  & \textbf{1.00}  & 0.97 & \textbf{1.00}  & 0.98 & 0.98\\
			\texttt{mra\_tra} & 0.99 & \textbf{1.00}  & 0.98 & \textbf{1.00}  & \textbf{1.00}  & \textbf{1.00}  & \textbf{1.00}  & \textbf{1.00}  & \textbf{1.00}  & \textbf{1.00}  & \textbf{1.00}  & \textbf{1.00}  & \textbf{1.00}  & \textbf{1.00}  & \textbf{1.00}  & \textbf{1.00}  & \textbf{1.00}  & \textbf{1.00}  & \textbf{1.00}  & \textbf{1.00} \\
			\texttt{payload} & 0.70 & \textbf{0.89}  & \textbf{0.91}  & 0.75 & 0.70 & 0.84 & 0.90 & 0.73 & 0.69 & \textbf{0.89}  & \textbf{0.91}  & 0.75 & 0.68 & 0.84 & 0.89 & 0.73 & 0.68 & 0.84 & 0.89 & 0.72\\
%			\texttt{proc} & 0.73 & \textbf{1.00}  & 0.97 & 0.82 & 0.70 & \textbf{1.00}  & 0.96 & 0.79 & 0.73 & \textbf{1.00}  & 0.96 & 0.81 & 0.73 & \textbf{1.00}  & 0.96 & 0.81 & 0.73 & \textbf{1.00}  & 0.97 & 0.82\\
			\bottomrule
	\end{tabular}}
\end{table}

\begin{figure}[t!]
\includegraphics[width=.55\textwidth]{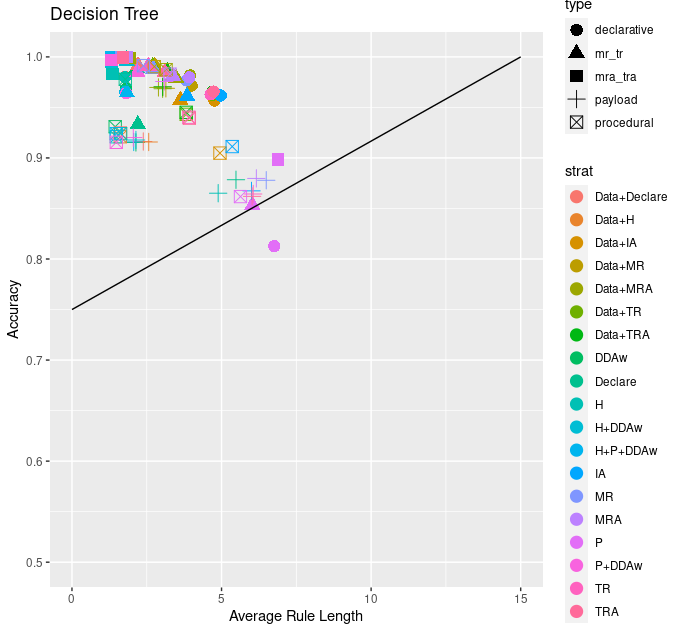}
\includegraphics[width=.55\textwidth]{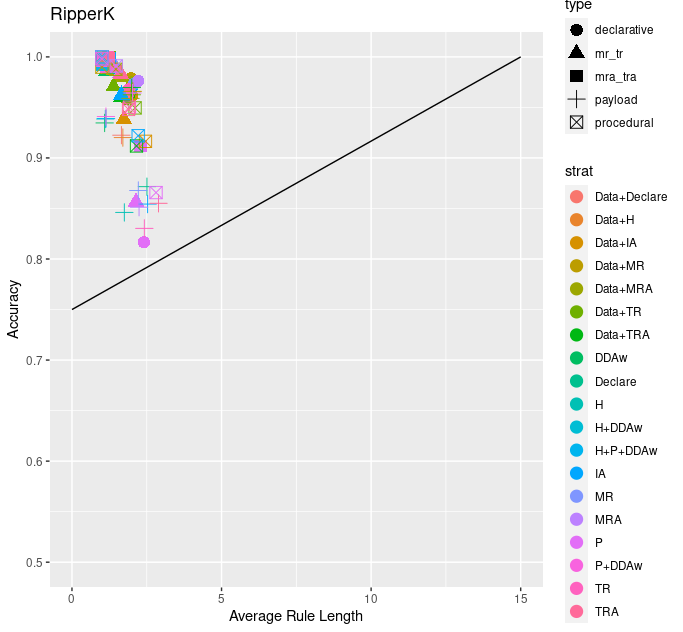}
\caption{Comparing Ripper$k$ and decision trees for both Average Rule Length and Accuracy.}\label{fig:DTRK}
%	\includegraphics[width=\textwidth]{figures/DT_1089_227.png}
%	\caption{decision tree: Differentiating rules by Accuracy and Average Rule Length for each hyper-parameter: $\{5,10,15,20,25,30,35,60,80\}$.}\label{fig:dtgoing}
	
%		\includegraphics[width=\textwidth]{figures/RK_1089_227.png}
%	\caption{Ripper$k$: Differentiating rules by Accuracy and Average Rule Length for each hyper-parameter: $\{0,2,4,6,8,10,16,18\}$.}\label{fig:rkgoing}
\end{figure}
%For the remainder of the experiments, we decided to plot only the results for accuracy: please remember that we still keep the same hyper-parameters as in the previous paragraph, so to establish whether there is a desirable correlation between precision and accuracy.

To answer research questions \textbf{RQ3} and \textbf{RQ4}, we compare the average length of the rules returned by Ripper$k$ and decision trees using all the feature encodings presented. In particular, in \figurename~\ref{fig:DTRK}, we plot the distribution of the precision of the mined rules and the average rule length. The diagonal black line represents the intuition that, by increasing the rule length, we would expect to have an increase in precision. As already remarked when answering the first two research questions, Ripper$k$ and decision trees have comparable accuracy, while the encodings combining data and control-flow have a better accuracy with respect to the ones based on data or control-flow only. However, Ripper$k$ provides the best trade-off of conciseness-accuracy since the distribution of its points is squeezed towards the y-axis with respect to the one obtained with decision trees. This effect is more evident when using encodings based on data or control-flow only.

\section{Conclusion}
\label{conclusions}

%{\color{red}[TODO: Rephrase on the light of the novel results]}

This paper has focused on approaches for uncovering and explaining (positive and negative) deviances in business process execution traces. In particular, three different aspects have been investigated.

First, different types of control-flow features were investigated (sequential, declarative and hybrid) for explaining deviances. The feature types were applied to different real-life logs, by showing advantages and limits of each of them. Overall, the conclusion was that hybrid encoding is preferable independently of the nature of the log and the labeling, provided that there is a real correlation between labels and control-flow. Note that this result is crucial in real settings in which the nature of the logs and the type of correlation features-labeling is not known a priori. 
%Indeed, using the hybrid encoding this correlation is discovered (if it exists) independently of the nature of the log and the labeling.

Second, the paper investigated the impact of data (together with declarative and sequential features) on the capability of existing techniques to explain process execution deviances. Data features were included using straightforward attribute extraction and aggregation methods and the discovery of data-aware {\declare} constraints. The results showed that the combination of data and control-flow features increases the performance of deviance mining. However, the extent of this improvement depends on the characteristics of the log and on the correlation between labels and features.
%The lack of improvement in performance can be due to a small or no correlation between the data and the labeling of a log.
%If there is no correlation between the data features and the labeling of a log, then the addition of the features will not help increasing the classification performance.

Third, this paper investigated the final outcomes of business process deviance mining returned to the user. More concretely, two different classifiers were evaluated (Ripper$k$ and decision trees) and compared them using accuracy metrics and in terms of conciseness of the returned decision rules. The results show that the accuracy of Ripper$k$ and decision trees are comparable, while Ripper$k$ returns shorter rules. In addition, Ripper$k$ is able to better exploit the availability of hybrid features with respect to decision trees in order to improve the accuracy of the classification. This result suggests that using Ripper$k$ as a classifier for business process deviance mining is a good alternative to decision trees, especially for providing more compact explanations to the user.

For future work, there are several possible directions to go and ways to improve the present work. One of the possible improvements would be to try out more feature selection methods to find better alternatives to the currently used \textit{coverage} method. To further improve the effectiveness of data features, more feature extraction methods could be considered, especially for extracting features based on meta-information. In order to assess the understandability of the returned rules, an empirical study could be carried out with human subjects for assessing whether more compact rules are actually better than longer ones. Another possible avenue for future work would be to experiment model-agnostic explainers for describing the important features in the classification process, which could allow the use of more complex and powerful classifiers with respect to the white-box classifiers used in this paper.

%In order to ease the finding of the best possible explanations, the hyperparameter search should be done automatically (e.g. search of coverage, maximum tree depths) to get the best explanations for a desired metric. For improvement to research in business process deviance mining, the creation of more concrete benchmarks for comparison of approaches is important.

%Another possible direction would be to try out model-agnostic methods for describing the important features in classification process, which could allow the use of more complex and powerful models. 

%\section*{Acknowledgement}
%This work was supported by the Estonian Research Council (grant IUT20-55)

\bibliographystyle{elsarticle-num}
\bibliography{thesis}

\end{document}